\documentclass[a4paper,10pt]{article}
\usepackage[utf8]{inputenc}

\usepackage{amsmath, amsfonts, amssymb}
\usepackage{amsthm, thmtools}     % for theorems and \theoremstyle
\usepackage[mathscr]{eucal}
\usepackage{xfrac}

\usepackage{geometry}
\usepackage{cleveref}
\usepackage{autonum}
\usepackage{color}
\usepackage{enumerate}
\usepackage[title]{appendix}

\usepackage[round]{natbib}

\usepackage{pgfplots}

 \allowdisplaybreaks
 
\newcommand{\T}{\ensuremath{\mathsf{T}}}

\newcommand{\tensor}[1]{\boldsymbol{\mathscr{#1}}}

\newcommand{\normf}[1]{\|{#1}\|_{\sf F}}

\newcommand{\norm}[1]{\left\lVert#1\right\rVert}
\newcommand{\abs}[1]{\left|#1\right|}

\newcommand{\scalar}[2]{\left\langle#1, #2\right\rangle}
\newcommand{\E}[1]{\mathbb{E}\left\{#1\right\}}
\newcommand{\pr}[1]{\mathbb{P}\left\{#1\right\}}

\newcommand{\convas}{\ensuremath{\overset{\text{a.s.}}{\longrightarrow}}}

\newcommand{\lambc}{\lambda_{\text{c}}}
\newcommand{\lambs}{\lambda_{\text{s}}}

\newcommand{\xml}{\hat{x}_{\sf ML}}

\DeclareMathOperator*{\argmax}{arg\,max}

\DeclareMathOperator{\tr}{tr}

\newtheorem{dfn}{Definition}
\newtheorem{thrm}[dfn]{Theorem}
\newtheorem{prp}[dfn]{Proposition}
\newtheorem{lem}[dfn]{Lemma}
\newtheorem{conjec}[dfn]{Conjecture}
\declaretheorem[style=remark,sibling=dfn]{remark}

\title{A Random Matrix Perspective on Random Tensors}
\author{José Henrique de Morais Goulart,$^1$ Romain Couillet$^2$ and Pierre Comon$^2$ \\[4mm]
    \normalsize 1: IRIT, Université de Toulouse, Toulouse INP, 31071 Toulouse, France \\  
    \normalsize\tt henrique.goulart@irit.fr \\[2mm]
    \normalsize 2: Université Grenoble Alpes, CNRS, Grenoble INP, GIPSA-lab, 38000 Grenoble, France \\ 
    \normalsize\tt \{romain.couillet, pierre.comon\}@gipsa-lab.grenoble-inp.fr
   }

\begin{document}

\maketitle

\begin{abstract}
Tensor models play an increasingly prominent role in many fields, notably in machine learning. In several applications, such as community detection, topic modeling and Gaussian mixture learning, one must estimate a low-rank signal from a noisy tensor. Hence, understanding the fundamental limits of estimators of that signal inevitably calls for the study of random tensors. Substantial progress has been recently achieved on this subject in the large-dimensional limit. Yet, some of the most significant among these results--in particular, a precise characterization of the abrupt phase transition (with respect to signal-to-noise ratio) that governs the performance of the maximum likelihood (ML) estimator of a symmetric rank-one model with Gaussian noise--were derived based of mean-field spin glass theory, which is not easily accessible to non-experts.

In this work, we develop a sharply distinct and more elementary approach, relying on standard but powerful tools brought by years of advances in random matrix theory. The key idea is to study the spectra of random matrices arising from contractions of a given random tensor. We show how this gives access to spectral properties of the random tensor itself. For the aforementioned rank-one model, our technique yields a hitherto unknown fixed-point equation whose solution precisely matches the asymptotic performance of the ML estimator above the phase transition threshold in the third-order case. A numerical verification provides evidence that the same holds for orders 4 and 5, leading us to conjecture that, for any order, our fixed-point equation is equivalent to the known characterization of the ML estimation performance that had been obtained by relying on spin glasses. Moreover, our approach sheds light on certain properties of the ML problem landscape in large dimensions and can be extended to other models, such as asymmetric and non-Gaussian.

\end{abstract}

\section{Introduction}

Discovering and exploiting underlying low-dimensional structures in data are key to the success of modern machine learning and signal processing techniques. 
In many applications, prior knowledge on the process or generative model that produces the observations leads to a low-rank tensor model, which postulates that the sought information takes the form of a tensor decomposition having a few algebraically simple terms---examples can be found in \citep{AnanHKT-14-JMLR,SidiDFH-17-TSP,Moru-11-DMKD,Land12}.
While this approach often yields satisfying results thanks to the the strong uniqueness properties enjoyed by such models \citep{SidiDFH-17-TSP}, which ensure their identifiability (and therefore their interpretability), the performance of tensor methods is quite difficult to anticipate and very few of them---which are specialized to restrictive models of rather narrow applicability---are actually accompanied by performance guarantees \citep{GeHJY-15-JMLR,JainO-14-NIPS,HuanMGW-15-PJO,AnanHKT-14-JMLR}.
This difficulty stems in large part from the notorious and significant conceptual challenges that accompany tensor models, as many results and notions pertaining to matrices do not readily generalize to higher-order tensors.

Recent years have seen some important progress in understanding the properties of low-rank tensor models of the form $\tensor{Y} = \tensor{X} + \tensor{W}$, with $\tensor{W}$ standing for additive noise and $\tensor{X}$ having the so-called canonical polyadic decomposition (CPD) form \citep{Hitc-27-JMP}, \citep{Land12}
\begin{equation}
\label{CPD}
     \tensor{X} = \sum_{r=1}^R \, \lambda_r \, a^{(1)}_r \otimes \dots \otimes a^{(d)}_r,
\end{equation}
where the natural number $R$ defines the rank of the model, $a^{(i)}_r \in \mathbb{R}^{N_i}$ and $\lambda_r$ is a positive number for $i=1,\ldots,d$ and $r=1,\ldots,R$. 
In particular, the performance of low-rank CPD estimators under various priors for the vectors $a^{(i)}_r$ is now better understood thanks to the efforts of several authors \citep{RichM-14-NIPS,PerrWB-20-AIHP,LesiMLK-17-ISIT,ChenHL-18-arXiv,JagaLM-20-AAP}.
This progress has been achieved by adopting a large-dimensional inferential statistics viewpoint, which amounts to studying the inference of the random variables $\lambda_r,  \, a^{(i)}_r$ from noisy observations (distributed according to some known law), under the crucial assumption that the dimensions grow large, that is, $N_i \rightarrow \infty$. Just like happens for random matrices, the latter assumption allows exploiting certain concentration-of-measure phenomena and leads to quite precise results, some of which are briefly reviewed below.

Despite these recent results bringing a substantial advancement in the understanding of random tensor models, they are still limited in scope and have in large part been derived by heavily relying on spin glass theory, thus being quite difficult to understand for non-expert readers. 
Also, their extension to more structured (and more relevant for applications) models is likely difficult with this approach.

We take a completely different path in this work, relying upon years of advances in the field of random matrix theory (RMT) in order to study random tensor models. 
The key idea lies in the study of \emph{contractions} of a given tensor model with unit norm vectors, which give rise to a set of random matrices. 
In particular, we show that with a judicious choice of such contractions one gains access to spectral properties of a random tensor in the sense introduced by \cite{Lim-05-CAMSAP}, which in turn allow a characterization of the performance of the maximum likelihood (ML) estimator of the rank-one symmetric spiked tensor model. 
For this model, our approach produces precise predictions matching recent results that had been previously obtained only through sophisticated ideas borrowed from spin glass theory by \cite{JagaLM-20-AAP}. 
Our findings are obtained by invoking the Gaussian tools developed by \cite{PastS-11-book}, and can be extended to other models, notably with non-Gaussian noise. 
This approach amounts to a significant methodological contribution, as it opens the door to the investigation of other, more general tensor models by means of random matrix tools,  on which a vast and mature body of work has been established in recent decades  \citep{PastS-11-book, BaiS-10-book, Tao-12-book}.

As such, the present work follows after the latest breakthroughs in the understanding, analysis and improvement of machine learning methods based on spiked large-dimensional random matrix models. Recent examples in data classification include the large-dimensional analysis of support vector machines \citep{liao2019large,kammoun2021precise}, semi-supervised and transfer learning \citep{mai2018random,mai2021consistent,tiomoko2020deciphering}, logistic regression \citep{mai2019large}, as well as neural network dynamics \citep{advani2020high,liao2018dynamics}. Similar ideas in data and graph clustering are found in \citep{couillet2016kernel,nadakuditi2012graph}.

\paragraph{Statistical performance of rank-one spike estimation.}

Our approach will be leveraged to study ML estimation of a planted signal vector $x$ in a rank-one symmetric model of the form
\begin{equation}
\label{SRO}
     \tensor{Y} =  \lambda \, x^{\otimes d} + \frac{1}{\sqrt{N}} \tensor{W},
\end{equation}
with $x$ uniformly distributed on the unit sphere $\mathbb{S}^{N-1}$ (which is known as a spherical prior) and $\tensor{W}$ a symmetric Gaussian noise tensor (see Section \ref{sec:GOE} for a precise definition). 
This is the so-called \emph{spiked rank-one tensor model}, whose ``signal'' part is referred to as a spike, with the positive number $\lambda$ effecively playing the role of a signal-to-noise ratio parameter.

Similarly to other high-dimensional statistical inference problems, the fundamental asymptotic performance limits of estimators of $x$ are of utmost importance in this context. Given some estimator $\hat{x}(\tensor{Y})$ taking values on $\mathbb{S}^{N-1}$, a natural performance measure is the mean \emph{alignment} (or overlap), defined as 
\begin{equation}
  \bar{\alpha}_{d,N}(\lambda) := \E{\abs{\scalar{x}{\hat{x}(\tensor{Y})}}} \qquad \in [0,1].
\end{equation} 
Two crucial questions regarding the behavior of this quantity as $N \rightarrow \infty$ are as follows:
\begin{enumerate}
 \item (\emph{Weak recovery}) For which range of $\lambda$ is there an estimator such that $\limsup_{N \rightarrow \infty} \bar{\alpha}_{d,N}(\lambda) > 0$?

 \item (\emph{Best asymptotic alignment}) What is the largest value of $\limsup_{N \rightarrow \infty} \bar{\alpha}_{d,N}(\lambda)$ attainnable by some estimator for a given value of $\lambda$?
 
\end{enumerate}
A clear understanding of these questions has been achieved in recent years.
As it turns out, there exists an $O(1)$ (in $N$) statistical threshold $\lambc(d)$ (which can be numerically computed) such that weak recovery is impossible when $\lambda < \lambc(d)$, but is possible when $\lambda > \lambc(d)$. Furthermore, for any $\lambda$ the ML estimator of $x$, which boils down to computing the best symmetric rank-one approximation of $\tensor{Y}$ or, equivalently, to solving
\begin{equation}
\label{ML}
  \max_{u \in \mathbb{S}^{N-1}} 
    \scalar{\tensor{Y}}{u^{\otimes d}} = 
  \max_{u \in \mathbb{S}^{N-1}} 
    \sum_{i_1 =1}^N \dots  \sum_{i_d = 1}^N \, Y_{i_1 \dots i_d} \ u_{i_1} \dots u_{i_d},
\end{equation}
attains the highest possible correlation among all measurable estimators of $x$ \citep{JagaLM-20-AAP}.

Before summarizing our own contributions, we will briefly review existing studies on this topic, and more generally on the estimation (and detection) of spiked tensor models with other priors.

\paragraph{Related work.}

The first study that seems to have looked into the spiked tensor model \eqref{SRO} was carried out by \cite{RichM-14-NIPS}, who showed that weak recovery can be performed by ML estimation whenever $\lambda$ exceeds a number $\mu_0(d)$, defined as the almost sure limit of $\frac{1}{\sqrt{N}} \, \norm{\tensor{W}}$, where $\norm{\tensor{W}}$ stands for the spectral norm of $\tensor{W}$ (see Section \ref{sec:tens} for precise definitions).
Then, \cite{MontRZ-17-TIT} investigated the closely related problem of detecting the presence of a rank-one symmetric spike given a realization of $\tensor{Y}$, which amounts to distinguishing the hypothesis $\lambda = 0$ from $\lambda \neq 0$. 
By using the second-moment method, they proved that the law of $\tensor{Y}$ is asymptotically indistinguishable\footnote{More precisely, the total variation distance between these laws converges to zero as $N \rightarrow \infty$.} from that of $\frac{1}{\sqrt{N}}\tensor{W}$ when $\lambda$ is smaller than a certain value $\lambda_{\sf 2nd}(d) < \mu_0(d)$ (for which a variational characterization was given), and thus no hypothesis test can possibly perform this task better than pure random guessing.
An extension of this result to a rank-$R$ spiked tensor was later derived by \cite{ChevL-18-WSSP}.

\cite{PerrWB-20-AIHP} observed that detection is also possible (asymptotically) when $\lambda >  \mu_0(d)$, since $\pr{\norm{\tensor{Y}}  > \mu_0(d)} = 1 - o(1)$ holds in this regime.
Interestingly, they found that this already holds for $\lambda$ larger than a certain bound strictly smaller than $\mu_0(d)$, beyond which the spike can thus be detected, and also (weakly) estimated. 
This effect is reminiscent of the so-called BBP transition of spiked random matrix models, named after the authors that first exposed it \citep{BaikBP-05-AP}; see also \citep{BenaN-11-AiM} for a quite general treatment of this phenomenon.
By developing a modified second-order method, \cite{PerrWB-20-AIHP} also found that detection and weak estimation are impossible when $\lambda$ is below some threshold $\lambda'_{\sf 2nd}(d) > \lambda_{\sf 2nd}(d)$, thus also improving upon the previously existing lower bound on the statistical threshold. Moreover, they derived similar results for Rademacher and sparse Rademacher priors on $x$.
Borrowing ideas from spin glass theory, \cite{Chen-19-AS} refined the results of \cite{PerrWB-20-AIHP} for the Rademacher prior, computing a single critical value that separates the regimes where the rank-one spike is detectable (in the sense that the total variation distance between the laws of $\tensor{Y}$ and of $\frac{1}{\sqrt{N}}\tensor{W}$ converges to 1) and can be weakly recovered, and where $\tensor{Y}$ and $\frac{1}{\sqrt{N}}\tensor{W}$ are statistically indistinguishable (with the aforementioned distance converging to zero) and the spike cannot be weakly recovered.

Also on the basis of results pertaining to mean-field spin glass models, \cite{ChenHL-19-AAP} derived a variational expression for the threshold that applies to both detection and estimation of a rank-one symmetric spiked tensor model with Gaussian noise. However, their model differs from \eqref{SRO} in that the vector $x$ has i.i.d.~elements distributed according to some prior.
Specifically, the estimation threshold is defined in that work as the value beyond which the minimum mean-squared error (MMSE) estimator performs strictly better (in the MSE sense) than any so-called ``dummy'' estimator (that does not depend on the data), and below which it performs as well as a dummy estimator.
They went on to extend this result to the case with multiple spikes (that is, higher rank) with possibly different priors: essentially, the signal-to-noise parameter $\lambda_r$ of at least one spike must exceed the threshold associated with its prior, or else neither detection nor estimation can possibly be performed. 

For the rank-one symmetric setting of \eqref{SRO} with spherical prior and Gaussian noise, \cite{JagaLM-20-AAP} finally settled the search for a precise expression for the phase transition threshold $\lambc(d)$ that separates the regime where one can detect and weakly estimate $x$ ($\lambda > \lambc(d)$) from the one where these tasks are impossible ($\lambda < \lambc(d)$), again by relying substantially upon results coming from spin glass theory.
They showed that this transition is much more abrupt for tensors of order $d > 2$ than for matrices, by giving also an exact asymptotic expression for the alignment of $x$ with the ML solution $\xml(\tensor{Y})$, namely $|\langle \xml(\tensor{Y}), x \rangle|$, which displays a jump discontinuity at $\lambc(d)$ (see Section \ref{sec:phasetran} for a precise description of that behavior, shown in Figure \ref{fig:align-spnorm}). 
Their analysis also produced an exact expression for the spectral norm of $\tensor{Y}$, which is very much in line with the BPP-like phenomenon reported by \cite{PerrWB-20-AIHP}, and showed that the ML estimator is information-theoretically optimal, in the sense that its alignment with $x$ matches that of the Bayes-optimal estimator.

Another interesting related work has been carried out by \cite{ArouMMN-19-CPAM}, who studied the complexity of the maximum likelihood landscape by using the Kac-Rice formula to compute the expected number of critical points and of local maxima of \eqref{ML}.  
Among several other conclusions, their study shows that for $\lambda < \lambc(d)$ the objective function values of all local maxima (including the global one) tend to be concentrated (as $N \rightarrow \infty$) on a quite narrow interval, whereas for $\lambda > \lambc(d)$ the value achieved by the global maximum ``escapes'' that interval and grows with $\lambda$.

\paragraph{Summary of contributions.}

Our main contribution is of a methodological flavor, and is based on studying random matrices obtained by \emph{contractions of a random tensor model}.
The fruitfulness of this idea is shown in the case of the rank-one symmetric spiked tensor model $\tensor{Y}$ given by \eqref{SRO}.
In this setting, the ML estimator $\xml(\tensor{Y})$ of the planted vector $x$ is shown to be the dominant eigenvector of the matrix arising from the $(d-2)$-fold contraction of $\tensor{Y}$ with $\xml(\tensor{Y})$ itself.
This is a simple consequence of the tensor eigenvalue equations (see Section \ref{sec:MLE-eigen}), which by definition characterize the critical points of \eqref{ML}.
By pursuing this connection, and under the assumption that there exists a sequence of critical points $(\mu, u)$ of \eqref{ML} such that $\abs{\scalar{x}{u}}$ and $\mu$ converge almost surely to deterministic functions of $\lambda$, we are able to deploy tools from matrix theory in order to establish a fixed-point characterization of these deterministic quantities.
As we will discuss, it turns out that for $d \in \{3,4,5\}$ the solution to our fixed-point equation precisely matches the results of \cite{JagaLM-20-AAP} describing the behavior of the ML estimator (and thus of global maxima of \eqref{ML}) beyond the critical value $\lambc(d)$. We conjecture that this actually holds for any $d \ge 3$.

In the process of deriving these results, we demonstrate that the spectral measure of contractions of $\tensor{Y}$ with sequences of independent vectors or of local maximizers of the ML problem (satisfying a mild technical condition)
converges weakly almost surely to a semi-circle law supported on $[-\beta_d, \beta_d]$, where $\beta_d = 2/\sqrt{d(d-1)}$.
This sheds light on one feature of the maximum likelihood landscape that had been exposed by \cite{ArouMMN-19-CPAM}, namely, the fact that the value of the objective function of \eqref{ML} at every local maxima asymptotically lies on the interval $[2\beta_3, \norm{\tensor{Y}}]$ when $d=3$. 
It turns out that every such a maximum is the largest eigenvalue of a contraction of $\tensor{Y}$, and the second-order necessary optimality conditions combined with the derived spectral measure of such contractions entail the lower bound $(d-1) \beta_d = 2 \beta_3$, while the upper bound follows by definition of $\norm{\tensor{Y}}$. 

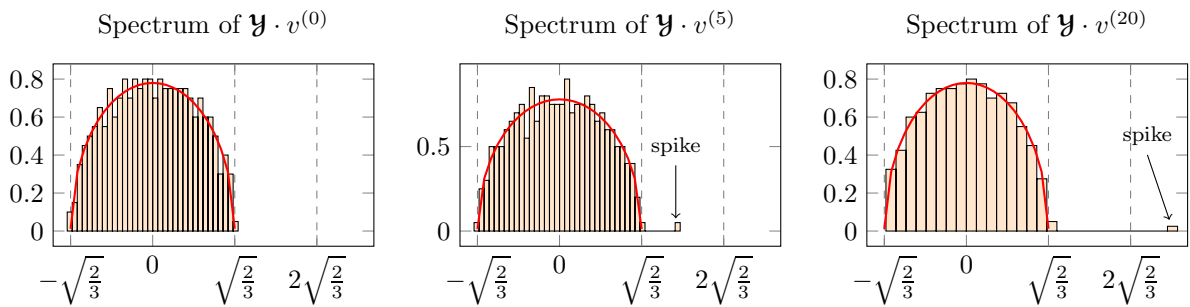
\begin{figure}[t]
    \centering
    \begin{tikzpicture}
\begin{axis}[
width=5.8cm,
height=4cm,
% xticklabel interval boundaries, 
% xticklabel={$[\pgfmathprintnumber\tick,\pgfmathprintnumber\nexttick)$}, 
xmajorgrids={false}, 
xmin=-0.99,
xmax=2.2,
title={Spectrum of $\tensor{Y} \cdot v^{(0)}$},
xtick={0},
extra x ticks={-0.8164,0.8164,1.633},
extra x tick labels={$-\sqrt{\frac{2}{3}}$,$\sqrt{\frac{2}{3}}$,$2\sqrt{\frac{2}{3}}$},
extra tick style={grid=major, grid style={dashed, gray}},
]
    \addplot[ybar interval, fill = white!80!orange]
        table[row sep={\\}]
        {
            \\
            -0.85  0.10000000000000014  \\
            -0.8  0.15000000000000022  \\
            -0.75  0.3500000000000005  \\
            -0.7  0.4500000000000006  \\
            -0.65  0.5000000000000007  \\
            -0.6  0.5500000000000008  \\
            -0.55  0.6500000000000009  \\
            -0.5  0.5500000000000008  \\
            -0.45  0.7500000000000011  \\
            -0.4  0.6000000000000009  \\
            -0.35  0.700000000000001  \\
            -0.3  0.8000000000000012  \\
            -0.25  0.700000000000001  \\
            -0.2  0.8000000000000012  \\
            -0.15  0.7500000000000011  \\
            -0.1  0.8000000000000012  \\
            -0.05  0.8000000000000012  \\
            0.0  0.700000000000001  \\
            0.05  0.8000000000000012  \\
            0.1  0.7500000000000011  \\
            0.15  0.7500000000000011  \\
            0.2  0.7500000000000011  \\
            0.25  0.7500000000000011  \\
            0.3  0.7500000000000011  \\
            0.35  0.700000000000001  \\
            0.4  0.6000000000000009  \\
            0.45  0.700000000000001  \\
            0.5  0.6000000000000009  \\
            0.55  0.6000000000000009  \\
            0.6  0.5000000000000007  \\
            0.65  0.30000000000000043  \\
            0.7  0.4000000000000006  \\
            0.75  0.30000000000000043  \\
            0.8  0.05000000000000007  \\
            0.85  0.0  \\
        }
        ;
\addplot+[no markers,domain=-0.8164:0.8164,color=red,thick] {3*sqrt(2/3-x^2)/(pi)};

\end{axis}
\end{tikzpicture}
\hspace{2mm}
\begin{tikzpicture}
\begin{axis}[
width=5.8cm,
height=4cm,
% xticklabel interval boundaries, 
% xticklabel={$[\pgfmathprintnumber\tick,\pgfmathprintnumber\nexttick)$}, 
xmajorgrids={false}, 
xmin=-0.99,
xmax=2.2,
title={Spectrum of $\tensor{Y} \cdot v^{(5)}$},
xtick={0},
extra x ticks={-0.8164,0.8164,1.633},
extra x tick labels={$-\sqrt{\frac{2}{3}}$,$\sqrt{\frac{2}{3}}$,$2\sqrt{\frac{2}{3}}$},
extra tick style={grid=major, grid style={dashed, gray}},
]
    \addplot[ybar interval, fill = white!80!orange]
        table[row sep={\\}]
        {
            \\
            -0.85  0.05000000000000007  \\
            -0.8  0.25000000000000033  \\
            -0.75  0.30000000000000043  \\
            -0.7  0.5000000000000007  \\
            -0.65  0.5000000000000007  \\
            -0.6  0.5000000000000007  \\
            -0.55  0.6000000000000009  \\
            -0.5  0.6500000000000009  \\
            -0.45  0.700000000000001  \\
            -0.4  0.7500000000000011  \\
            -0.35  0.5500000000000008  \\
            -0.3  0.8500000000000012  \\
            -0.25  0.6500000000000009  \\
            -0.2  0.8000000000000012  \\
            -0.15  0.8000000000000012  \\
            -0.1  0.7500000000000011  \\
            -0.05  0.7500000000000011  \\
            0.0  0.7500000000000011  \\
            0.05  0.9000000000000012  \\
            0.1  0.700000000000001  \\
            0.15  0.7500000000000011  \\
            0.2  0.7500000000000011  \\
            0.25  0.8000000000000012  \\
            0.3  0.7500000000000011  \\
            0.35  0.6500000000000009  \\
            0.4  0.700000000000001  \\
            0.45  0.6000000000000009  \\
            0.5  0.6000000000000009  \\
            0.55  0.5000000000000007  \\
            0.6  0.5000000000000007  \\
            0.65  0.4000000000000006  \\
            0.7  0.4000000000000006  \\
            0.75  0.2000000000000003  \\
            0.8  0.05000000000000007  \\
            0.85  0.0  \\
            0.9  0.0  \\
            0.95  0.0  \\
            1.0  0.0  \\
            1.05  0.0  \\
            1.1  0.0  \\
            1.15  0.05000000000000007  \\
            1.2  0.0  \\
        }
        ;
\addplot+[no markers,domain=-0.8164:0.8164,color=red,thick] {3*sqrt(2/3-x^2)/(pi)};

\node (p1) at (axis cs:1.156,0.5)  {\footnotesize spike};
\node (linep1) at (axis cs:1.156, 0.02) {}; 
\draw [->] (p1.south) -- (linep1);

\end{axis}
\end{tikzpicture}
\hspace{2mm}
\begin{tikzpicture}
\begin{axis}[
width=5.8cm,
height=4cm,
% xticklabel interval boundaries, 
% xticklabel={$[\pgfmathprintnumber\tick,\pgfmathprintnumber\nexttick)$}, 
xmajorgrids={false}, 
xmin=-0.99,
xmax=2.2,
title={Spectrum of $\tensor{Y} \cdot v^{(20)}$},
xtick={0},
extra x ticks={-0.8164,0.8164,1.633},
extra x tick labels={$-\sqrt{\frac{2}{3}}$,$\sqrt{\frac{2}{3}}$,$2\sqrt{\frac{2}{3}}$},
extra tick style={grid=major, grid style={dashed, gray}},
]
    \addplot[ybar interval, fill = white!80!orange]
        table[row sep={\\}]
        {
            \\
            -0.8  0.32499999999999973  \\
            -0.7  0.4249999999999996  \\
            -0.6  0.5999999999999994  \\
            -0.5  0.6249999999999994  \\
            -0.4  0.7249999999999993  \\
            -0.3  0.7499999999999993  \\
            -0.2  0.7499999999999993  \\
            -0.1  0.7749999999999994  \\
            0.0  0.7999999999999993  \\
            0.1  0.7749999999999994  \\
            0.2  0.6999999999999994  \\
            0.3  0.7249999999999993  \\
            0.4  0.6749999999999994  \\
            0.5  0.5499999999999995  \\
            0.6  0.4499999999999996  \\
            0.7  0.27499999999999974  \\
            0.8  0.049999999999999954  \\
            0.9  0.0  \\
            1.0  0.0  \\
            1.1  0.0  \\
            1.2  0.0  \\
            1.3  0.0  \\
            1.4  0.0  \\
            1.5  0.0  \\
            1.6  0.0  \\
            1.7  0.0  \\
            1.8  0.0  \\
            1.9  0.0  \\
            2.0  0.024999999999999977  \\
            2.1  0.0  \\
        }
        ;
\addplot+[no markers,domain=-0.8164:0.8164,color=red,thick] {3*sqrt(2/3-x^2)/(pi)};

\node (p1) at (axis cs:1.8,0.5)  {\footnotesize spike};
\node (linep1) at (axis cs:2.05, 0.01) {}; 
\draw [->] (p1.south) -- (linep1);

\end{axis}
\end{tikzpicture}
    \caption{Tensor power iteration method applied to find a local maximum of the ML problem \eqref{ML}, with $d=3$ and $N=500$. This algorithm produces a sequence $v^{(k)}$ which (under certain conditions) converges to a tensor eigenvector $u$ associated with a local maximum of \eqref{ML} \citep{KoldM-11-SIMAX}.}
    \label{fig:power}
\end{figure}

Our analysis also offers another perspective into the behavior of the tensor power iteration method \citep{KoldM-11-SIMAX}, which can be seen to produce a sequence of \emph{matrices}\footnote{Here, the notation $\tensor{Y} \cdot (v^{(k)})^{d-2}$ stands for the $(d-2)$-fold contraction of $\tensor{Y}$ with the vector $v^{(k)}$ (see Section \ref{sec:tens} for our notational conventions).} $\tensor{Y} \cdot (v^{(k)})^{d-2}$ converging (under appropriate conditions) to a contraction of $\tensor{Y}$ with one of its eigenvectors $u$ (in the sense introduced by \cite{Lim-05-CAMSAP}), as $v^{(k)} \rightarrow u$.
In this interpretation, each iteration is tantamount to a single step of a \emph{matrix} power iteration method applied to $\tensor{Y} \cdot (v^{(k)})^{d-2}$, which then changes in the next iteration.
In particular, when this algorithm is applied to a spiked tensor model of the form \eqref{SRO}, the spectra of such matrices $\tensor{Y} \cdot (v^{(k)})^{d-2}$ exhibit a ``bulk'' obeying a semi-circle law and a spike that moves away from that bulk until it reaches the position of an eigenvalue associated with a local maximum (see Figure \ref{fig:power} for an illustration), which asymptotically must be at least as large as $(d-1)\beta_d$.
The presence of this spike in $\tensor{Y} \cdot (v^{(k)})^{d-2}$ is observed regardless of the value of $\lambda$, and is in fact a consequence of the dependence between $\tensor{Y}$ and $v^{(k)}$.

Finally, the high degree of versatility of our approach should be emphasized, as it can be extended to other, less stringent models with other noise distributions and asymmetric structure.

\vskip4mm

We point out that this work was first presented at the seminar in Random Tensors of Texas A\&M University \citep{GoulCC-21-RTseminar} and at the 2021 SIAM Conference on Applied Linear Algebra \citep{GoulCC-21-SIAMLA}.

%%%%%%%%%%%%%%%%%%%%%%%%%%%%%%%%%%%%%%%%%%%%%%%%%%%%%%%%%%%%%%%%%%%%%%%%%%%%%%%%%%%%%%%%%%%%%%%%%%%%
%%%%%%%%%%%%%%%%%%%%%%%%%%%%%%%%%%%%%%%%%%%%%%%%%%%%%%%%%%%%%%%%%%%%%%%%%%%%%%%%%%%%%%%%%%%%%%%%%%%%
%%%%%%%%%%%%%%%%%%%%%%%%%%%%%%%%%%%%%%%%%%%%%%%%%%%%%%%%%%%%%%%%%%%%%%%%%%%%%%%%%%%%%%%%%%%%%%%%%%%%

\section{Preliminaries and main results}

We will start by introducing some notation and defining the objects of interest in our study, before describing our main results.
For simplicity, in the following we shall stick to symmetric random tensors, but the main ideas generalize to the asymmetric case.

% ================================================================================================================
% ================================================================================================================

\subsection{Tensors and contractions}
\label{sec:tens}

The entries of a $d$-th order tensor $\tensor{Y} \in \mathbb{R}^{N \times \dots \times N}$ will be denoted by uppercase letters, as in $Y_{i_1 \, \dots \, i_d}$. The same convention applies to matrices ($d=2$) but not to vectors, whose components are written in lowercase letters, as in $u_i$.
A tensor is said to be symmetric if and only if its entries are invariant with respect to any permutation of the indices, that is, $Y_{i_1 \, \dots \, i_d} = Y_{j_1 \, \dots \, j_d}$, where $(j_1,\ldots,j_d)$ is an arbitrary permutation of $(i_1,\ldots,i_d)$.
The space of $d$-th order real-valued symmetric tensors of dimensions $N \times \dots \times N$ will be denoted $\mathcal{S}^d(N)$.

Given a tensor $\tensor{Y} \in \mathcal{S}^d(N)$ and vectors $u^{(1)}, \ldots, u^{(p)} \in \mathbb{R}^N$ with $p \le d$, we define the contraction $\tensor{Y} \cdot (u^{(1)},\ldots,u^{(p)})$ as the $(d-p)$th-order tensor with components
\begin{equation}
% \label{def-con}
\left(  \tensor{Y} \cdot (u^{(1)},\ldots,u^{(p)}) \right)_{i_{p+1}, \ldots, i_d}
 =
 \sum_{i_1=1}^N \dots \sum_{i_p=1}^N \, Y_{i_1 \, \dots \, i_d} \, u^{(1)}_{i_1} \, \dots \, u^{(p)}_{i_p}.
\end{equation} 
The order of the vectors in $(u^{(1)},\ldots,u^{(p)})$ is immaterial, due to the symmetry of $\tensor{Y}$. 
Moreover, the outcome is clearly symmetric as well, by symmetry of $\tensor{Y}$.
In particular, when $p = d-2$, $p = d-1$ and $p = d$ we get a symmetric $N \times N$ matrix, an $N$-dimensional vector and a scalar, respectively.
A special case of interest to us is when $u^{(j)}=u$ for all $j \in \{1,\ldots,p\}$, for which we introduce the notation
\begin{equation}
 \label{def-con-sym}
\tensor{Y} \cdot u^{p} := \tensor{Y} \cdot (\underbrace{u,\ldots,u}_{p \text{ times}}).
\end{equation} 
In a similar vein, the multilinear transformation of $\tensor{Y} \in \mathcal{S}^d(N)$ by a square matrix $U \in \mathbb{R}^{N \times N}$ is defined as
\begin{equation}
\label{mult-tr}
  \tensor{Y}' = \tensor{Y} \cdot U^d 
\quad \Leftrightarrow \quad
  Y'_{i_1 \, \ldots \, i_d} =  
 \sum_{n_1=1}^N \dots \sum_{n_d=1}^N \, Y_{n_1 \, \dots \, n_d} \, u_{i_1 n_1} \, \dots \, u_{i_d n_d}.
\end{equation} 
When $U$ is nonsingular, this operation amounts to a change of basis in $\mathcal{S}^d(N)$.

A $d$-th order tensor $\tensor{X} \in \mathbb{R}^{N \times \dots \times N}$ is said to be of rank one if and only there exist vectors $u^{(1)}, \ldots, u^{(d)} \in \mathbb{R}^N$ such that the entries of $\tensor{X}$ decompose as $X_{i_1 \, \dots \, i_d} = u^{(1)}_{i_1} \, \dots \, u^{(d)}_{i_d}$. This relation can be expressed in terms of the tensor product $\otimes$ as follows:
\[
   \tensor{X} = u^{(1)} \otimes \dots \otimes u^{(d)}.
\]
In particular, a symmetric rank-one tensor can be written in this form with $u^{(i)} = u$ for all $i$, in which case we denote it as $u^{\otimes d}$. A tensor is said to have rank $R$ if it can be written as a sum of $R$ rank-one terms, but not fewer. Similarly, a symmetric tensor has symmetric rank $S$ if and only if it can be written as a sum of $S$ symmetric rank-one terms, but not fewer.

By equipping $\mathbb{R}^{N \times \dots \times N}$ with the Euclidean scalar product \citep{Hack12} 
\begin{equation}
 \langle \tensor{Y}, \tensor{X} \rangle := \sum_{i_1=1}^N \dots \sum_{i_d=1}^N \,
   Y_{i_1 \, \dots \, i_d} \, X_{i_1 \, \dots \, i_d},
\end{equation} 
we have the identity $\scalar{\tensor{Y}}{u^{\otimes d}} = \tensor{Y} \cdot u^d$ for any symmetric rank-one $d$th-order tensor $u^{\otimes d}$, and also  $\scalar{u^{\otimes d}}{v^{\otimes d}} = \scalar{u}{v}^d$ for any two such tensors. Naturally, the Euclidean or Frobenius norm on $\mathbb{R}^{N \times \dots \times N}$ is defined as $\normf{\tensor{Y}} := \sqrt{\scalar{\tensor{Y}}{\tensor{Y}}}$.

% ================================================================================================================
% ================================================================================================================

\subsection{Random tensors and the Gaussian orthogonal tensor ensemble}
\label{sec:GOE}

A random tensor $\tensor{Y}$ is simply a tensor-valued random variable with an associated probability measure $\rho_{\tensor{Y}}$. 

In particular, the Gaussian orthogonal matrix ensemble can be readily generalized to tensors in $\mathcal{S}^d(N)$ by defining the density
\begin{equation}
\label{GOE}
    f(\tensor{W}) = \frac{1}{Z_d(N)} \, e^{-\frac{1}{2} \normf{ \tensor{W}}^2},
\end{equation}
where $Z_d(N)$ is a normalization constant. Just as in the matrix case, because $f$ depends on $\tensor{W}$ only through its Euclidean norm, it is invariant with respect to an orthogonal change of basis of $\mathcal{S}^d(N)$, as
 \[
  \forall \, U \in \text{O}(N), \quad \normf{ \tensor{W} \cdot U^d} = \normf{\tensor{W}},
\]
where $\text{O}(N)$ denotes the orthogonal group. Another consequence of the definition \eqref{GOE} is that, by symmetry of $\tensor{W}$, the variance of an entry $W_{i_1 \dots  i_d}$ equals the reciprocal of the number of distinct permutations of its indices $(i_1, \ldots, i_d)$. For instance, in the $d=3$ case one can expand $\normf{ \tensor{W} }^2$ as 
\begin{equation}
\normf{ \tensor{W} }^2 = 
    \sum_{i} W^2_{i  i  i} +
    3 \sum_{i \neq j} W^2_{i  i  j} + 
    6 \sum_{i < j < k} W^2_{i  j  k},
\end{equation}
which shows that on-diagonal entries of $\tensor{W}$ have variance 1, entries with two distinct indices have variance 1/3 and entries with three distinct indices have variance 1/6:
\begin{equation}
\label{var-GOE}
    \sigma^2_{W_{i  j  k}} = 
    \frac{1}{6} + \delta_{ij} \frac{1}{6} + \delta_{ik}\frac{1}{6}
                           + \delta_{jk}\frac{1}{6} + \delta_{ij} \delta_{jk}\frac{1}{3},
\end{equation}
where $\delta_{ij}$ equals $1$ if $i = j$ and $0$ otherwise.

% ================================================================================================================
% ================================================================================================================

\subsection{The contraction ensemble of a random tensor}

A commonly used expedient when dealing with tensors is to consider matrix slices or matrix unfoldings. 
A slice is obtained by fixing all but two indices of a tensor, which yields a matrix. For instance, one can define the map $S_{\tensor{Y}} : \{1,\ldots,N\}^{d-2} \rightarrow \mathbb{R}^{N \times N}$ by requiring that the components of the matrix $S_{\tensor{Y}}(n_1, \ldots, n_{d-2})$ satisfy
\[
 \left( S_{\tensor{Y}}(n_1, \ldots, n_{d-2}) \right)_{i, j} =  Y_{n_1, \ldots, n_{d-2}, i, j}.
\]
Note that the order of the indices on the right-hand side (and of the arguments as well) is immaterial, as $\tensor{Y}$ is symmetric. Hence, this map produces an object of dimension lower than that of $\tensor{Y}$.
By contrast, an unfolding (also known as flattening or matricization) of a tensor accommodates all its entries inside a matrix by partitioning its set of indices into two subsets, which are then bijectively mapped into row and column indices.
These two definitions are very useful as they enable the application of matrix-analytic concepts and tools to tensors, and have long been studied in algebraic geometry \citep{Land12} and utilized in various scientific domains for data analysis purposes \citep{Tuck-66-Psy, BroWMK-97-ASR}. 
In particular, a spectral method for the estimation of the spike in the rank-one model \eqref{SRO} based on unfoldings has been proposed by \cite{RichM-14-NIPS}, and later studied by \cite{ArouHH-21-arXiv}.

When it comes to the study of random tensors, one is tempted to resort to these definitions in order to borrow tools from the well-developed RMT.
However, merely taking slices or unfoldings of a tensor turns out to be insufficient for our purposes, as these matrices cannot describe the whole picture and in particular do not provide access to the \emph{spectral} properties of the tensor (for us here its $\ell_2$-eigenpairs, which are connected to the ML estimation problem as we will shortly discuss).
Nevertheless, realizing that the above defined map $S_{\tensor{Y}}$ amounts to the contraction of $\tensor{Y}$ with unit norm vectors 
\[
S_{\tensor{Y}}(n_1, \ldots, n_{d-2}) = \tensor{Y} \cdot (e^{(n_1)}, \ldots, e^{(n_{d-2})}),
\quad 
\text{where} 
\quad 
e^{(n)}_i = \delta_{in},
\]
one can more generally consider arbitrary contractions of a random tensor with unit-norm vectors. 
For our purposes, though, it will be sufficient to restrict these contractions to the case where all vectors are the same. This motivates the following definition.

\begin{dfn}[Contraction ensemble]
\label{dfn:con}
Let $\tensor{Y} \in \mathcal{S}^d(N)$ be a random tensor. We define the \emph{contraction ensemble} of $\tensor{Y}$ as the set of random matrices 
\begin{equation}
    \mathcal{M}(\tensor{Y})
    :=
    \Bigg\{ \tensor{Y} \cdot v^{d-2} \, : \, v \in \mathbb{S}^{N-1} \Bigg\}.
\end{equation}
\end{dfn}

As will be shown ahead in Section \ref{sec:ROSTM}, by studying appropriately chosen random matrices from this set one gains access to spectral properties of $\tensor{Y}$.

\subsection{Main results}

We are now ready to describe our main results.

\subsubsection{Spectral measure of contractions of the Gaussian orthogonal tensor ensemble}
\label{sec:spec-meas}

Our first result concerns the spectral measure of random matrices from the contraction ensemble $\mathcal{M}\left(\frac{1}{\sqrt{N}}\tensor{W}\right)$ of a random tensor $\tensor{W}$ distributed according to \eqref{GOE}. This result is of its own interest, as taking contractions of a Gaussian symmetric tensor with indpendent entries (apart from the symmetries) is a natural way of introducing dependencies in a random matrix model.\footnote{Indeed, according to the authors of \citep{AuG-21-arXiv}, the first draft of our work motivated them to study the spectral asymptotics of contractions of tensors of a Wigner type. They argued that existing methods for the study of random matrices with dependent entries do not apply in this case, and then proposed a graph formalism for dealing with it on the basis of combinatorial arguments. In its turn, our proof is based on analytic tools and focused on Gaussian tensors, but can be extended to other distributions by means of the tools from \cite{PastS-11-book}.}
Furthermore, it plays a major role in our analysis of the ML problem \eqref{ML}. 

\begin{thrm}
\label{thm:spec-meas}
Let $\tensor{W} \in \mathcal{S}^d(N)$ be a sequence of random tensors with density given by \eqref{GOE}, with $d \ge 3$. For any  deterministic sequence\footnote{To keep the notation simple, we adopt the usual practice of omitting the index of the sequences of vectors $v$ and of random tensors $\tensor{W}$. This convention is adopted throughout the paper.} of vectors $v \in \mathbb{S}^{N-1}$, the empirical spectral measure of $\frac{1}{\sqrt{N}}\tensor{W} \cdot v^{d-2}$ converges weakly almost surely to the semi-circle distribution whose Stieljes transform reads 
\begin{equation}
   m_d(z) =  \frac{2}{\beta_d^2} \left( -z + z \sqrt{ 1 - \frac{\beta_d^2}{z^2} } \right),
\end{equation} 
with $\beta_d := 2/\sqrt{d(d-1)}$. Its density is thus given by
\begin{equation}
 \label{sclaw-W}
    \rho(dx) = \frac{2}{\pi \, \beta_d^2} \, \sqrt{\left(\beta_d^2 - x^2\right)^+} \, dx,
\end{equation}
and is supported on $[-\beta_d, \beta_d]$.
\end{thrm}

The proof of this result can be essentially obtained as a byproduct of that of Theorem \ref{thm:main}.
Concretely, it will turn out that the limiting spectral measure of the contractions $\tensor{W} \cdot v^{d-2}$ with deterministic $v \in \mathbb{S}^{N-1}$ is exactly the same as that of $\tensor{Y} \cdot u^{d-2}$, where the vectors $u$ form a sequence of critical points $u$ of the ML problem satisfying certain conditions explained ahead.
We comment on how the proof of Theorem \ref{thm:spec-meas} can be derived from that of Theorem \ref{thm:main} in Section \ref{sec:spec-meas}.

\begin{figure}
    \centering
    \begin{tikzpicture}
\begin{axis}[
width=8cm,
height=4cm,
% xticklabel interval boundaries, 
% xticklabel={$[\pgfmathprintnumber\tick,\pgfmathprintnumber\nexttick)$}, 
xmajorgrids={false}, 
ymax = 1.35,
xmin=-1,
xmax=1,
title={$d=3$, $N=500$},
extra x ticks={-0.8164,0.8164},
extra x tick labels={$-\sqrt{\frac{2}{3}}$,$\sqrt{\frac{2}{3}}$},
extra tick style={grid=major, grid style={dashed, gray}},
]
    \addplot[ybar interval, fill = white!80!orange]
        table[row sep={\\}]
        {
            \\
            -0.85  0.04000000000000005  \\
            -0.8  0.2400000000000003  \\
            -0.75  0.3600000000000005  \\
            -0.7  0.44000000000000056  \\
            -0.65  0.5200000000000007  \\
            -0.6  0.5200000000000007  \\
            -0.55  0.6000000000000008  \\
            -0.5  0.6400000000000008  \\
            -0.45  0.6800000000000008  \\
            -0.4  0.720000000000001  \\
            -0.35  0.6800000000000008  \\
            -0.3  0.800000000000001  \\
            -0.25  0.720000000000001  \\
            -0.2  0.720000000000001  \\
            -0.15  0.760000000000001  \\
            -0.1  0.800000000000001  \\
            -0.05  0.760000000000001  \\
            0.0  0.8400000000000011  \\
            0.05  0.760000000000001  \\
            0.1  0.760000000000001  \\
            0.15  0.760000000000001  \\
            0.2  0.720000000000001  \\
            0.25  0.800000000000001  \\
            0.3  0.6800000000000008  \\
            0.35  0.6400000000000008  \\
            0.4  0.6000000000000008  \\
            0.45  0.720000000000001  \\
            0.5  0.5600000000000007  \\
            0.55  0.5600000000000007  \\
            0.6  0.5200000000000007  \\
            0.65  0.44000000000000056  \\
            0.7  0.3200000000000004  \\
            0.75  0.28000000000000036  \\
            0.8  0.04000000000000005  \\
            0.85  0.0  \\
        }
        ;
\addplot+[no markers,samples=50,domain=-0.8164:0.8164,color=red,thick] {3*sqrt(2/3-x^2)/(pi)};

\end{axis}
\end{tikzpicture}
\hspace{1cm}
\begin{tikzpicture}
\begin{axis}[
width=8cm,
height=4cm,
% xticklabel interval boundaries, 
% xticklabel={$[\pgfmathprintnumber\tick,\pgfmathprintnumber\nexttick)$}, 
xmajorgrids={false}, 
ymax = 1.35,
xmin=-1,
xmax=1,
title={$d=4$, $N=200$},
xtick={-1,0,1},
extra x ticks={-0.5773,0.5773},
extra x tick labels={$-\sqrt{\frac{1}{3}}$,$\sqrt{\frac{1}{3}}$},
extra tick style={grid=major, grid style={dashed, gray}},
]
    \addplot[ybar interval, fill = white!80!orange]
        table[row sep={\\}]
        {
            \\
            -0.65  0.09999999999999991  \\
            -0.6  0.09999999999999991  \\
            -0.55  0.39999999999999963  \\
            -0.5  0.5999999999999994  \\
            -0.45  0.6999999999999994  \\
            -0.4  0.7999999999999993  \\
            -0.35  0.9999999999999991  \\
            -0.3  1.099999999999999  \\
            -0.25  0.8999999999999992  \\
            -0.2  1.099999999999999  \\
            -0.15  0.8999999999999992  \\
            -0.1  1.099999999999999  \\
            -0.05  1.099999999999999  \\
            0.0  1.299999999999999  \\
            0.05  1.099999999999999  \\
            0.1  1.099999999999999  \\
            0.15  0.8999999999999992  \\
            0.2  0.9999999999999991  \\
            0.25  0.9999999999999991  \\
            0.3  0.8999999999999992  \\
            0.35  0.7999999999999993  \\
            0.4  0.7999999999999993  \\
            0.45  0.6999999999999994  \\
            0.5  0.2999999999999997  \\
            0.55  0.09999999999999991  \\
            0.6  0.09999999999999991  \\
            % 0.65  0.0  \\
        }
        ;
\addplot+[no markers,samples=50,domain=-0.5773:0.5773,color=red,thick] {6*sqrt(1/3-x^2)/(pi)};

\end{axis}
\end{tikzpicture}
    \caption{Density of the spectral measure of Theorem \ref{thm:spec-meas} (in red) for $d=3,4$ and histogram of eigenvalues for one realization of $\frac{1}{\sqrt{N}} \, \tensor{W} \cdot v$, with a random vector $v$ drawn uniformly on $\mathbb{S}^{N-1}$ and independently of $\tensor{W}$.}
    \label{fig:spmeas}
\end{figure}
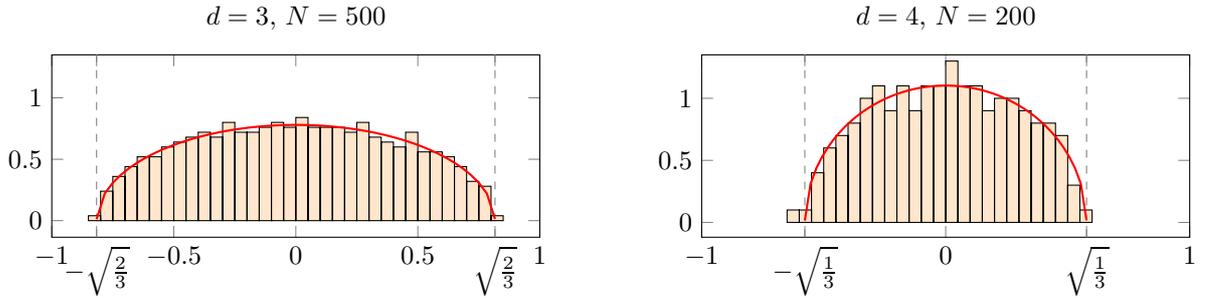

Though the statement of Theorem \ref{thm:spec-meas} might at first glance seem trivial since $\tensor{W} \cdot v^{d-2}$ is a linear combination of Gaussian matrices, these matrices are \emph{not} independent due to the symmetries in $\tensor{W}$.
Nevertheless, at the end we show that, just as for standard Wigner matrices, the limiting Stieltjes transform also satisfies a quadratic equation, as the additional terms due to the dependence structure all vanish asymptotically in the derivation.
An illustration of the derived spectral measure is shown on Figure \ref{fig:spmeas}.

One heuristic argument to see why the size of the support should scale like $O(1/d)$ is as follows. Take every vector $v$ in the considered sequence to be the first canonical vector of the standard basis, denoted by $e^{(1)}$. Then, the contraction $\frac{1}{\sqrt{N}}\tensor{W} \cdot v^{d-2} = \frac{1}{\sqrt{N}}\tensor{W} \cdot (e^{(1)})^{d-2}$ yields a matrix $C$ with entries 
\[
   C_{ij} = \frac{1}{\sqrt{N}} W_{1 \dots 1 i j}.
\]
It follows from the definition of $\tensor{W}$ that $C$ is by construction a Gaussian Wigner matrix whose entries have variance equal to the reciprocal of the number of distinct permutations of $(1,\ldots,1,i,j)$. With the exception of the first row and first column of $C$, the variances are therefore $\frac{1}{d(d-1)}$ off-diagonal and $\frac{2}{d(d-1)}$ on the main diagonal. Hence, apart from the first row and column (whose contribution vanishes as $N \rightarrow \infty$), we see that $C$ equals a standard GOE matrix (with variance 1 off-diagonal and 2 on the diagonal) multiplied by $\frac{1}{\sqrt{d(d-1)}}$, and hence its limiting spectrum should be a semicircle with support\footnote{This is the same result found by \cite{AuG-21-arXiv} for Wigner-type tensors when all $d-2$ contractions are with the same vector. By contrast, when the contractions are allowed to involve $d-2$ distinct vectors, the support size can shrink down to $O(1/\sqrt{d!})$---this can be understood along the lines of our heuristic argument by considering contractions with the canonical basis vectors $e^{(1)}, \ldots, e^{(d-2)}$.} $[-\beta_d, \beta_d]$. 

\subsubsection{Fixed-point characterization of the ML estimator}

We now state our main result concerning the performance of the ML estimator of model \eqref{SRO}. In order to derive it, we study a sequence of contractions $\tensor{Y} \cdot u^{d-2}$, where the vectors $u$ are critical points of the ML problem satisfying some conditions discussed below.
The result is as follows.

\begin{thrm}
\label{thm:main}
Let $d \ge 3$, fix $\lambda > 0$ and suppose that there exists a sequence of critical points $u$ of the maximum likelihood problem \eqref{ML} such that 
\begin{align}
 \label{conv-align}
  \scalar{x}{u} \convas & \ \alpha_{d,\infty}(\lambda),    \\
  \tensor{Y} \cdot u^d \convas & \ \mu_{d,\infty}(\lambda) 
  \label{conv-mu}
\end{align} 
for some deterministic functions $\alpha_{d,\infty}(\lambda)$ and $\mu_{d,\infty}(\lambda)$ defined on $\lambda$. Furthermore, assume that the following inequalities hold: $\alpha_{d,\infty}(\lambda) > 0$ and $\mu_{d,\infty}(\lambda) > (d-1) \beta_d$. Then, $\mu_{d,\infty}(\lambda)$ satisfies the fixed-point equation $\mu_{d,\infty}(\lambda) = \phi_d(\mu_{d,\infty}(\lambda), \lambda)$, where
\begin{align}
\label{fp-1}
 \phi_d(z, \lambda) := & \ \lambda \, \omega_d^d(z, \lambda) - \frac{1}{d-1} \, 
                              m_d\left( \frac{z}{d-1} \right), \\
 \omega_d(z, \lambda) := & \ \left[ \frac{1}{\lambda} \left( z + \frac{1}{d}\,  m_d\left( \frac{z}{d-1} \right) \right) \right]^{\frac{1}{d-2}}.
\label{fp-2}
\end{align} 
Furthermore, $\alpha_{d,\infty}(\lambda) = \omega_d(\mu_{d,\infty}(\lambda), \lambda)$.
\end{thrm}

For $d=3$, we are able to verify that this fixed-point equation leads to the same exact expressions that follow from the study of \cite{JagaLM-20-AAP} for the asymptotic spectral norm of $\tensor{Y}$ (given by $\mu_{d,\infty}$) and the alignment $|\scalar{x}{\xml(\lambda)}|$ of the ML estimator (given by $\alpha_{d,\infty}(\lambda)$). These expressions are as follows.

\begin{lem}
\label{lem:fpeq}
For $d=3$, the only positive solution to the fixed-point equation of Theorem \ref{thm:main}  for positive values of $\lambda$ is given by
\begin{equation}
    \mu_{3,\infty}(\lambda) = \frac{3\lambda^2 + \lambda\sqrt{9\lambda^2 - 12}+4}{\sqrt{18\lambda^2 + 6\lambda\sqrt{9\lambda^2 - 12}}},
\end{equation}
which holds for $\lambda \ge 2/\sqrt{3}$. Furthermore, 
\begin{equation}
    \alpha_{3,\infty}(\lambda) = \sqrt{\frac{1}{2} + \sqrt{\frac{3\lambda^2 - 4}{12 \lambda^2}}}.
\end{equation}
\end{lem}

Unfortunately, showing that this holds true for any $d \ge 3$ is likely to be difficult, and we do not pursue this path here. 
In particular, for $d \ge 6$ the variational characterization of the alignment given by \cite{JagaLM-20-AAP} does not admit an explicit expression, and thus another strategy would be needed for such a proof.
Nonetheless, for $d = 4$ and $d=5$, explicit expressions for both quantities (spectral norm and alignment) do exist, and we verify numerically that they satisfy the equations \eqref{fp-1}--\eqref{fp-2}. We thus conjecture that the same happens for every $d \ge 3$.

Before moving on, let us comment on the assumptions of Theorem \ref{thm:main}. 
The condition $\mu_{d,\infty}(\lambda) > (d-1) \beta_d$ is essentially imposed for a technical reason, related to the use of the implicit function theorem when computing derivatives of $u$ and $\mu$ with respect to $\tensor{W}$. 
As we will argue ahead, it is not restrictive for our purposes, since every local maximum $u$ of the ML problem (including global ones) is such that $\tensor{Y} \cdot u^d \ge (d-1) \beta_d$ with probability $1 - o(1)$. 
The condition $\alpha_{d,\infty}(\lambda) > 0$, in its turn, is required at the end of our asymptotic analysis for deriving equations \eqref{fp-1}--\eqref{fp-2}.
Since $\alpha_{d,\infty}(\lambda)$ is the limiting alignment $\scalar{x}{u}$, this condition means that the considered sequence cannot converge to any critical point (asymptotically) lying on the orthogonal complement of $x$. 
Therefore, it prevents us from ``seeing'' the ML solution $\xml$ in the regime where it is asymptotically uncorrelated with $x$, that is, below the critical value $\lambc(d)$. 

Finally, the assumption that there exists a sequence of critical points such that both quantities in \eqref{conv-align}--\eqref{conv-mu} converge almost surely is of course a strong one, and is crucial to our analysis. 
At present, it is not clear how one can rigorously show that such a sequence exists in the framework of our approach: it seems hard to unambiguously pin down a sequence of critical points for which standard arguments (namely, a moment control combined with Markov's inequality and Borel-Cantelli) can be applied to show almost sure convergence.
Our result indicates that, for $\lambda > \lambc(d)$, such a sequence asymptotically ``behaves'' like a global maximum, in the sense that its alignment and cost function value match (for $d=3,4,5$) what is known for the ML estimator in that regime.
However, the only characterization that we have at our disposal corresponds to the tensor eigenvalue equations, which hold by definition for all critical points, combined with the condition $\mu_{d,\infty}(\lambda) > (d-1) \beta_d$, due to which one could probably restrict the sequence to contain local maxima only.

As mentioned before, the limiting spectral measure of the contractions $\tensor{Y} \cdot u^{d-2}$ with the sequence of critical points $u$ considered in Theorem \ref{thm:main} is the same as that described by Theorem \ref{thm:spec-meas}---indeed, the Stieltjes transform $m_d(z)$ is one of the main ingredients in the proof of Theorem \ref{thm:main}.
As we will see, it turns out that the additional terms introduced by the dependence of $u$ on $\tensor{W}$ (which are computed using the implicit function theorem and Gaussian integration by parts) all vanish asymptotically. 
Hence, the limiting spectrum of $\tensor{Y} \cdot u^{d-2}$ is still a semicircle law on $[-\beta_d, \beta_d]$.
(Moreover, if $u$ is a local maximum, then a spike can be seen in the spectrum of $\tensor{Y} \cdot u^{d-2}$ regardless of the value of $\lambda$, because $u$ is an eigenvector of that matrix with an eigenvalue larger than $(d-1) \beta_d$ with probability $1-o(1)$---see Figure \ref{fig:power} for an illustration).

%%%%%%%%%%%%%%%%%%%%%%%%%%%%%%%%%%%%%%%%%%%
%%%%%%%%%%%%%%%%%%%%%%%%%%%%%%%%%%%%%%%%%%%
%%%%%%%%%%%%%%%%%%%%%%%%%%%%%%%%%%%%%%%%%%%

\section{Maximum likelihood estimation of the symmetric rank-one spiked tensor model}
\label{sec:ROSTM}

In the following, we introduce the key ideas that underlie our analysis of the symmetric rank-one spiked tensor model \eqref{SRO}.
Since the spherical prior imposed on $x$ does not bring useful information for estimation purposes, for simplicity we will take the point of view in which $x$ is an arbitrary deterministic vector on the unit sphere.

\subsection{Maximum likelihood estimation, tensor and matrix eigenpairs}
\label{sec:MLE-eigen}

By writing the density of $\tensor{Y}$ as 
\begin{equation}
    f(\tensor{Y}) = \frac{1}{Z_d(N)} \, e^{-\frac{N}{2} \normf{ \tensor{Y} - \lambda \, x^{\otimes d} }^2},
\end{equation}
one can readily see that ML estimation of $x$ boils down to solving the best rank-one approximation problem
\begin{equation}
 \min_{u \in \mathbb{S}^{N-1}}
      \normf{ \tensor{Y} - \lambda \, u^{\otimes d} }^2.
\end{equation} 
Since $\lambda > 0$, it is straightforward to show that the above problem is equivalent to \eqref{ML}. For convenience, we will add a $\frac{1}{d}$ factor in the cost function of that problem, rewriting it as follows:
\begin{equation}
\label{def-f}
 \max_{u \in \mathbb{S}^{N-1}} \, f(u), 
 \qquad \text{where} \qquad f(u) := \frac{1}{d} \, \scalar{\tensor{Y}}{u^{\otimes d}} = \frac{1}{d} \, \tensor{Y} \cdot u^d.
\end{equation}

Now, writing the Lagrangian as $\mathcal{L}(u, \mu) := f(u) - \frac{\mu}{2}(\scalar{u}{u} - 1)$, it follows that stationary points $(\mu, u)$ must satisfy the Karush-Kuhn-Tucker conditions
\begin{equation}
 \frac{\partial \mathcal{L}}{\partial u}  = \nabla_f (u) - \mu u   = \tensor{Y} \cdot u^{d-1} - \mu u = 0
 \qquad \text{and} \qquad 
 \scalar{u}{u} = 1,
\end{equation} 
implying that they are (real-valued) solutions to the equations
\begin{equation}
\label{eig}
 \tensor{Y} \cdot u^{d-1} = \mu \, u, \qquad \scalar{u}{u} = 1.
\end{equation} 
These are called tensor $\ell_2$-eigenvalue\footnote{Other definitions of eigenpairs have been proposed for tensors. In the book on tensor spectral theory by \cite{QiZ-17-book}, solutions to \eqref{eig} are termed E-eigenpairs, and real-valued solutions are called Z-eigenpairs.} equations, following the variational definition of tensor eigenpairs introduced by \cite{Lim-05-CAMSAP}.
For simplicity, we will refer to $(\mu, u)$ satisfying \eqref{eig} as a tensor eigenpair, and to $\mu$ and $u$ as an eigenvalue and an eigenvector of $\tensor{Y}$, respectively. 
It should be borne in mind that, unlike the matrix case, the normalization $\norm{u} = 1$ is essential, since $\tensor{Y} \cdot (\alpha u)^{d-1} = \mu \, \alpha^{d-1} \, u \neq \mu \, \alpha \, u$ for $\alpha \neq 1$ and $d > 2$.

The above discussion shows that the spiked rank-one tensor model resembles in many respects its matrix counterpart, in that its ML estimator also corresponds to the eigenvector associated with the largest eigenvalue. 
However, there are several striking differences. 
One of them is computational, and has strong implications in practice: whereas this estimator can be computed in polynomial time in the matrix case, solving \eqref{ML} for $d > 2$ is worst-case NP-hard \citep{HillL-13-JACM}, and thus in general one can at best hope to find a local maximum of the nonconvex problem \eqref{def-f}.
Another difference concerns the sharp phase transition behavior that is observed in the tensor case, as we will discuss ahead.

Now, a simple but crucial observation that brings into focus the relevance of Definition \ref{dfn:con} is the following, whose proof follows trivially from the identity $\tensor{Y}\cdot u^{d-1} = (\tensor{Y} \cdot u^{d-2}) \, u$:

\begin{prp}
\label{prp:eig}
Let $\tensor{Y} \in \mathcal{S}^d(N)$ and $u \in \mathbb{S}^{N-1}$. The pair $(\mu, u)$ is an eigenpair of the tensor $\tensor{Y}$ if and only if it is an eigenpair of the matrix $\tensor{Y} \cdot u^{d-2} \in \mathcal{M}(\tensor{Y})$. 
\end{prp}

An immediate consequence is that, by a suitable choice of matrices from the contraction ensemble $\mathcal{M}(\tensor{Y})$, we can gain access to spectral properties of $\tensor{Y}$. We will thoroughly rely on this link to derive our results on the spiked symmetric rank-one tensor model.

\begin{remark}
\label{rem:mu}
Before moving on, we make another observation that will later prove useful. 
First note that, if $(\mu, u)$ is an eigenpair of $\tensor{Y}$, then it is also an eigenpair of the Hessian $\nabla^2_\mathcal{L} := \frac{\partial^2 \mathcal{L}}{\partial u^2}$ evaluated at $u$, since:
\begin{equation}
\label{eig-Hess}
 % \nabla^2_f(u) = (d-1) \, \tensor{Y} \cdot u^{d-2} \implies
 \nabla^2_\mathcal{L}(u) = (d-1) \, \tensor{Y} \cdot u^{d-2} - \mu \, I  
 \quad \implies \quad
 \nabla^2_\mathcal{L}(u) u = (d-1) \, \tensor{Y} \cdot u^{d-1} - \mu \, u = (d-2) \mu \, u.
\end{equation}
A necessary condition for $u$ to be a local maximum of problem \eqref{def-f} is that $\nabla^2_\mathcal{L}(u)$ be negative semidefinite on the tangent space at $u$ (see, e.g., Theorem 12.5 of \cite{NoceW-06-book}), that is, 
\begin{equation}
  \scalar{ \nabla^2_\mathcal{L}(u) \, w}{ w } \le 0,
  \qquad \forall \, w \in u^\perp := \{w \in \mathbb{R}^{N} \; | \; \scalar{u}{w} = 0\}.
\end{equation} 
Under the above condition, it follows from \eqref{eig-Hess} that 
\begin{equation}
\label{cond-mu}
 \max_{w \in \mathbb{S}^{N-1} \, \cap \, u^\perp} \scalar{ (\tensor{Y} \cdot u^{d-2}) \, w}{ w} \le \frac{\mu}{d-1}.
\end{equation}
This means that $\mu$ must be the largest eigenvalue of (the matrix) $\tensor{Y} \cdot u^{d-2}$, with all others bounded by $\mu/(d-1)$, which is true in particular for the largest eigenvalue of $\tensor{Y}$, as discussed above.
\end{remark}

Remark \ref{rem:mu} implies that, at every local maximum $u$ of the ML problem, the second largest eigenvalue of $\tensor{Y} \cdot u^{d-2}$ cannot be larger than $\mu/(d-1)$ and, consequently, the largest eigenvalue $\mu$ is isolated from the rest of the spectrum of $\tensor{Y} \cdot u^{d-2}$. 
At first sight this seems to be a particularly unsettling conclusion since, unlike the spiked Wigner random \emph{matrix} model, this implies the systematic presence of a ``spike'' in the spectrum of $\tensor{Y} \cdot u^{d-2}$ (positioned at least $d-1$ times as far as the bulk spectrum of $\tensor{Y} \cdot u^{d-2}$) irrespective of the (possibly null) signal-to-noise ratio $\lambda$.
Yet, as previously discussed, note that this spike can merely be induced by the contraction of $\tensor{W}$ with a dependent vector $u$, and thus may bear no relation whatsoever with the ``signal'' term  $\lambda \, x^{\otimes d}$.

% ================================================================

\subsection{Asymptotic performance of ML estimation}
\label{sec:phasetran}

Before stating our results, we briefly recall the contributions recently made by \cite{JagaLM-20-AAP}.
Let 
\[
  \varphi_d(\lambda, t) := \lambda^2 \, t^d + \log (1 - t) + t,
\]
and define $\lambc(d) := \sup\{ \lambda \ge 0 \, : \, \sup_{t \in [0,1[} \varphi_d(\lambda, t) \le 0\}$. In was shown in \citep{JagaLM-20-AAP} that the critical value $\lambc$ defines the phase transition point, below which ML estimation of the planted vector $x$ of the model \eqref{SRO} suddenly breaks down, in the sense that its solution (asymptotically) ceases to exhibit any correlation with $x$ as measured by the alignment $|\scalar{x}{\xml(\lambda)}|$.
More precisely, their result states that for $\lambda \neq \lambc(d)$,
\begin{equation}
    |\scalar{x}{\xml(\lambda)}|  \convas \sqrt{q_d(\lambda)}
\end{equation}
as $N \rightarrow \infty$, where 
\begin{equation}
\label{xTu-JagaML}
  q_d(\lambda) := \begin{cases}
                  0, & \text{if } \lambda < \lambc(d), \\ 
                  \argmax_{t \in [0,1[} \varphi_d(\lambda, t), & \text{if } \lambda > \lambc(d).
               \end{cases}
\end{equation}
Furthermore, the quantity $\mu^\star_d(\lambda) = \tensor{Y} \cdot \left(\xml(\lambda)\right)^d$ also converges almost surely to a deterministic expression given by
\begin{equation}
\label{mu-JagaML}
\mu^\star_d(\lambda) \convas
     \begin{cases}
     \mu_0(d),        & \lambda \le \lambc(d), \\
     \sqrt{d} \, \frac{1 + \lambda^2 q_d(\lambda)^{d-1}}{\sqrt{1 + \lambda^2 d q_d(\lambda)^{d-1}}}  & \lambda > \lambc(d),
 \end{cases}
\end{equation}
where $\mu_0(d) = (d + z^\star)/\sqrt{d + d z^\star}$, with $z^\star$ denoting the unique positive solution of 
\begin{equation}
  \frac{1 + z}{z^2}\log(1 + z) - \frac{1}{z} - \frac{1}{d} = 0.   
\end{equation}
As discussed in the Introduction, $\mu_0(d)$ is in fact the (almost surely) asymptotic value of $\frac{1}{\sqrt{N}}\|\tensor{W}\|$, and it can be shown that $\lambc(d) < \mu_0(d)$ for all $d$---hence the claim of a BBP-like phase transition taking place for all $d$. Similarly, $\mu^\star_d(\lambda)$ is in effect the asymptotic spectral norm of $\tensor{Y}$.

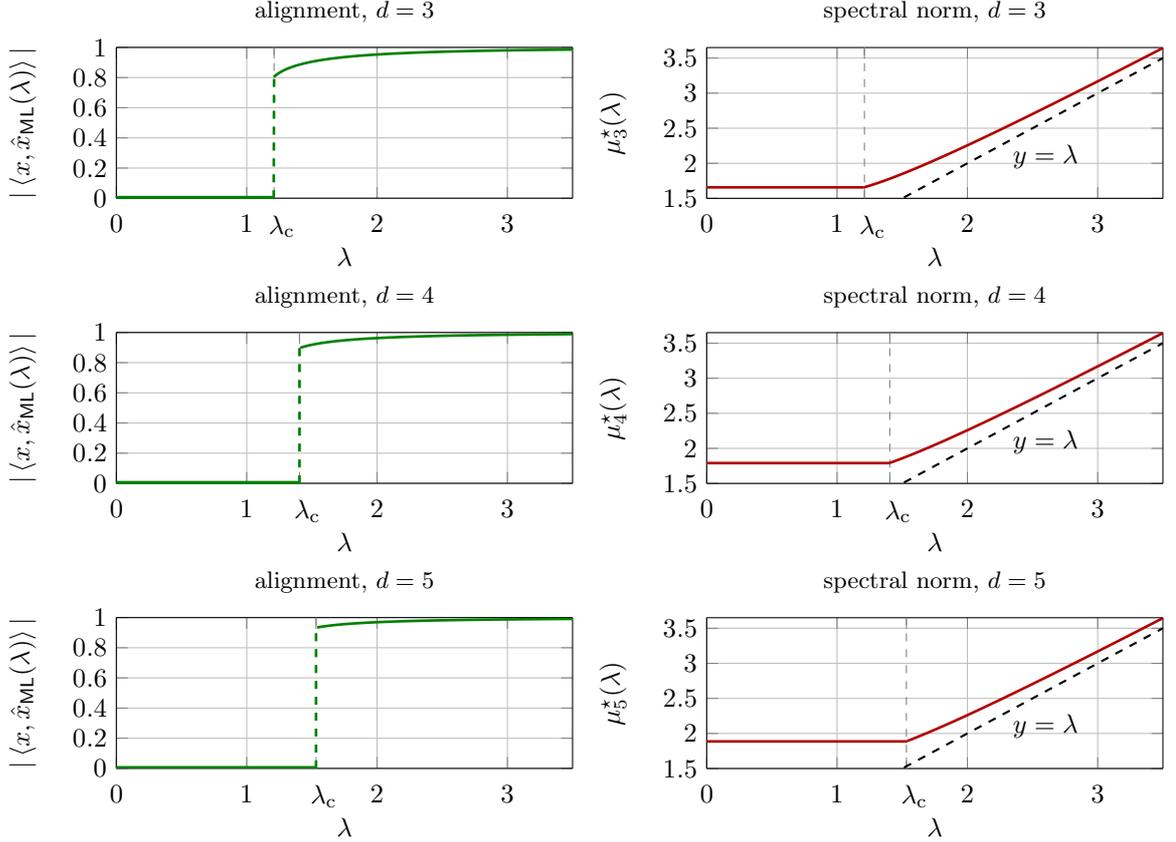
\begin{figure}[t!]
    \centering
    \begin{tikzpicture}

\begin{axis}[%
width=6cm,
height=2cm,
at={(0cm,0cm)},
scale only axis,
separate axis lines,
% every outer x axis line/.append style={black},
% every x tick label/.append style={font=\color{black}},
ymin=0,
ymax=1,
xlabel={$\lambda$},
ylabel={$|\scalar{x}{\xml(\lambda)}|$},
xmajorgrids,
% every outer y axis line/.append style={black},
% every y tick label/.append style={font=\color{black}},
xmin=0,
xmax=3.5,
extra x ticks={1.21},
extra x tick labels={\; $\lambc$},
extra tick style={grid=major, grid style={dashed, gray}},
ytick={0, 0.2, 0.4, 0.6, 0.8, 1},
ymajorgrids,
title style={font=\small},
legend style={at={(0.97,0.03)},anchor=south east,legend cell align=left,align=left,draw=black},
legend columns = 2,
xticklabel shift={.1cm},
legend style={font=\tiny},
yticklabel shift={.0cm},
title={alignment, $d=3$}
]

\addplot [color=green!50!black,solid,line width=1.8pt,forget plot]
  table[row sep=crcr]{%
            0.0  0.0  \\
            1.2  0.0  \\
};

\addplot+[no markers,domain=1.207:3.5,color=green!50!black,line width=1,solid,samples=100] {sqrt(1/2 + sqrt((3*x^2 - 4)/(12*x^2)))};

\addplot [color=green!50!black,dashed,line width=1pt,forget plot]
  table[row sep=crcr]{%
            1.207  0.0  \\
           1.21 0.8058535847162311 \\
};

\end{axis}
\end{tikzpicture}
    \begin{tikzpicture}

\begin{axis}[%
width=6cm,
height=2cm,
at={(0cm,0cm)},
scale only axis,
separate axis lines,
% every outer x axis line/.append style={black},
% every x tick label/.append style={font=\color{black}},
ymin=1.5,
ymax=3.65,
xlabel={$\lambda$},
ylabel={$\mu^\star_3(\lambda)$},
xmajorgrids,
% every outer y axis line/.append style={black},
% every y tick label/.append style={font=\color{black}},
xmin=0,
xmax=3.5,
extra x ticks={1.21},
extra x tick labels={\; $\lambc$},
extra tick style={grid=major, grid style={dashed, gray}},
ymajorgrids,
title style={font=\small},
legend style={at={(0.97,0.03)},anchor=south east,legend cell align=left,align=left,draw=black},
legend columns = 2,
xticklabel shift={.1cm},
legend style={font=\tiny},
yticklabel shift={.0cm},
title={spectral norm, $d=3$}
]

\addplot [color=red!70!black,solid,line width=1,forget plot]
  table[row sep=crcr]{%
            0.0  1.657000332100091  \\
            1.207  1.657000332100091  \\
};

\addplot+[no markers,domain=1.207:3.5,color=red!70!black,line width=1,solid,samples=100] {(3*x^2 + x*sqrt(9*x^2 - 12)+4)/sqrt(18*x^2 + 6*x*sqrt(9*x^2 - 12))};

\addplot+[no markers,dashed,domain=0:3.5,color=black,thick] {x};

% \addplot [color=darkorange,dashed,line width=1pt,forget plot]
%   table[row sep=crcr]{%
%             1.207  1.65  \\
%            1.21 3.65 \\
%};

 \node[] at (axis cs: 2.6,2.1) {$y = \lambda$};

\end{axis}
\end{tikzpicture}
    
    \begin{tikzpicture}

\begin{axis}[%
width=6cm,
height=2cm,
at={(0cm,0cm)},
scale only axis,
separate axis lines,
% every outer x axis line/.append style={black},
% every x tick label/.append style={font=\color{black}},
ymin=0,
ymax=1,
xlabel={$\lambda$},
ylabel={$|\scalar{x}{\xml(\lambda)}|$},
xmajorgrids,
% every outer y axis line/.append style={black},
% every y tick label/.append style={font=\color{black}},
xmin=0,
xmax=3.5,
extra x ticks={1.4050},
extra x tick labels={\; $\lambc$},
extra tick style={grid=major, grid style={dashed, gray}},
ytick={0, 0.2, 0.4, 0.6, 0.8, 1},
ymajorgrids,
title style={font=\small},
legend style={at={(0.97,0.03)},anchor=south east,legend cell align=left,align=left,draw=black},
legend columns = 2,
xticklabel shift={.1cm},
legend style={font=\tiny},
yticklabel shift={.0cm},
title={alignment, $d=4$}
]

\addplot [color=green!50!black,solid,line width=1.8pt,forget plot]
  table[row sep=crcr]{%
            0.0  0.0  \\
            1.4  0.0  \\
};

\addplot+[no markers,color=green!50!black,line width=1,solid] 
        table[row sep={\\}]
        {
            x  y  \\
            1.41  0.8982380995141006  \\
            1.42  0.9010686211082238  \\
            1.43  0.9037228786973027  \\
            1.44  0.9062211248079264  \\
            1.45  0.9085801042604866  \\
            1.46  0.9108138514046717  \\
            1.47  0.9129342680171951  \\
            1.48  0.9149515514758708  \\
            1.49  0.916874518113134  \\
            1.5  0.9187108515446987  \\
            1.51  0.9204672962416003  \\
            1.52  0.9221498104391259  \\
            1.53  0.9237636883762254  \\
            1.54  0.9253136590772901  \\
            1.55  0.926803966963837  \\
            1.56  0.9282384382288349  \\
            1.57  0.9296205359372732  \\
            1.58  0.9309534061131499  \\
            1.59  0.9322399165556544  \\
            1.6  0.9334826897420554  \\
            1.61  0.9346841308846638  \\
            1.62  0.9358464519884402  \\
            1.63  0.9369716925861448  \\
            1.64  0.9380617376963618  \\
            1.65  0.9391183334468375  \\
            1.66  0.9401431007244772  \\
            1.67  0.9411375471489496  \\
            1.68  0.94210307761535  \\
            1.69  0.9430410036099323  \\
            1.7  0.94395255146935  \\
            1.71  0.9448388697265084  \\
            1.72  0.9457010356637247  \\
            1.73  0.9465400611754624  \\
            1.74  0.9473568980276332  \\
            1.75  0.9481524425877775  \\
            1.76  0.9489275400898384  \\
            1.77  0.9496829884883489  \\
            1.78  0.9504195419493738  \\
            1.79  0.9511379140192142  \\
            1.8  0.9518387805065117  \\
            1.81  0.9525227821088199  \\
            1.82  0.9531905268107935  \\
            1.83  0.9538425920778054  \\
            1.84  0.954479526865907  \\
            1.85  0.9551018534665741  \\
            1.86  0.9557100692025167  \\
            1.87  0.9563046479889771  \\
            1.88  0.956886041773307  \\
            1.89  0.9574546818642061  \\
            1.9  0.9580109801607629  \\
            1.91  0.9585553302903559  \\
            1.92  0.9590881086635188  \\
            1.93  0.9596096754530352  \\
            1.94  0.9601203755037905  \\
            1.95  0.9606205391792483  \\
            1.96  0.9611104831498474  \\
            1.97  0.9615905111280935  \\
            1.98  0.9620609145546639  \\
            1.99  0.9625219732394428  \\
            2.0  0.9629739559610274  \\
            2.01  0.9634171210279335  \\
            2.02  0.9638517168044289  \\
            2.03  0.9642779822036649  \\
            2.04  0.9646961471505424  \\
            2.05  0.9651064330165343  \\
            2.06  0.9655090530285  \\
            2.07  0.9659042126533517  \\
            2.08  0.9662921099602808  \\
            2.09  0.9666729359621095  \\
            2.1  0.9670468749372051  \\
            2.11  0.9674141047332829  \\
            2.12  0.9677747970543144  \\
            2.13  0.9681291177316618  \\
            2.14  0.968477226980479  \\
            2.15  0.9688192796423318  \\
            2.16  0.9691554254149234  \\
            2.17  0.9694858090697436  \\
            2.18  0.9698105706583986  \\
            2.19  0.9701298457083243  \\
            2.2  0.9704437654085352  \\
            2.21  0.9707524567860126  \\
            2.22  0.9710560428732972  \\
            2.23  0.9713546428678045  \\
            2.24  0.9716483722833557  \\
            2.25  0.971937343094372  \\
            2.26  0.9722216638731593  \\
            2.27  0.9725014399206757  \\
            2.28  0.9727767733911491  \\
            2.29  0.9730477634108922  \\
            2.3  0.9733145061916328  \\
            2.31  0.9735770951386642  \\
            2.32  0.9738356209540923  \\
            2.33  0.9740901717354497  \\
            2.34  0.9743408330699184  \\
            2.35  0.9745876881243972  \\
            2.36  0.9748308177316287  \\
            2.37  0.9750703004725931  \\
            2.38  0.9753062127553588  \\
            2.39  0.9755386288905719  \\
            2.4  0.9757676211637536  \\
            2.41  0.9759932599045672  \\
            2.42  0.9762156135532031  \\
            2.43  0.9764347487240252  \\
            2.44  0.9766507302666135  \\
            2.45  0.9768636213243268  \\
            2.46  0.9770734833905066  \\
            2.47  0.9772803763624344  \\
            2.48  0.9774843585931478  \\
            2.49  0.9776854869412175  \\
            2.5  0.9778838168185785  \\
            2.51  0.9780794022365068  \\
            2.52  0.9782722958498252  \\
            2.53  0.9784625489994212  \\
            2.54  0.9786502117531496  \\
            2.55  0.9788353329451962  \\
            2.56  0.9790179602139684  \\
            2.57  0.9791981400385781  \\
            2.58  0.9793759177739799  \\
            2.59  0.9795513376848218  \\
            2.6  0.979724442978065  \\
            2.61  0.9798952758344244  \\
            2.62  0.9800638774386814  \\
            2.63  0.9802302880089161  \\
            2.64  0.9803945468247046  \\
            2.65  0.9805566922543234  \\
            2.66  0.9807167617810044  \\
            2.67  0.9808747920282765  \\
            2.68  0.9810308187844344  \\
            2.69  0.9811848770261675  \\
            2.7  0.9813370009413842  \\
            2.71  0.9814872239512638  \\
            2.72  0.9816355787315647  \\
            2.73  0.9817820972332232  \\
            2.74  0.9819268107022638  \\
            2.75  0.9820697496990555  \\
            2.76  0.9822109441169327  \\
            2.77  0.98235042320021  \\
            2.78  0.9824882155616123  \\
            2.79  0.9826243491991417  \\
            2.8  0.9827588515124043  \\
            2.81  0.9828917493184159  \\
            2.82  0.9830230688669075  \\
            2.83  0.9831528358551467  \\
            2.84  0.9832810754422955  \\
            2.85  0.9834078122633213  \\
            2.86  0.983533070442475  \\
            2.87  0.9836568736063548  \\
            2.88  0.9837792448965693  \\
            2.89  0.9839002069820142  \\
            2.9  0.9840197820707771  \\
            2.91  0.9841379919216828  \\
            2.92  0.9842548578554929  \\
            2.93  0.9843704007657704  \\
            2.94  0.9844846411294222  \\
            2.95  0.9845975990169303  \\
            2.96  0.9847092941022823  \\
            2.97  0.984819745672612  \\
            2.98  0.9849289726375587  \\
            2.99  0.9850369935383579  \\
            3.0  0.9851438265566674  \\
            3.01  0.9852494895231434  \\
            3.02  0.9853539999257702  \\
            3.03  0.9854573749179552  \\
            3.04  0.9855596313263949  \\
            3.05  0.9856607856587196  \\
            3.06  0.9857608541109256  \\
            3.07  0.9858598525745997  \\
            3.08  0.9859577966439446  \\
            3.09  0.9860547016226103  \\
            3.1  0.9861505825303392  \\
            3.11  0.986245454109428  \\
            3.12  0.9863393308310165  \\
            3.13  0.9864322269012054  \\
            3.14  0.9865241562670085  \\
            3.15  0.9866151326221482  \\
            3.16  0.9867051694126934  \\
            3.17  0.9867942798425504  \\
            3.18  0.9868824768788076  \\
            3.19  0.9869697732569396  \\
            3.2  0.9870561814858757  \\
            3.21  0.9871417138529358  \\
            3.22  0.9872263824286386  \\
            3.23  0.9873101990713856  \\
            3.24  0.9873931754320247  \\
            3.25  0.9874753229582962  \\
            3.26  0.9875566528991669  \\
            3.27  0.9876371763090536  \\
            3.28  0.9877169040519385  \\
            3.29  0.9877958468053833  \\
            3.3  0.987874015064442  \\
            3.31  0.9879514191454741  \\
            3.32  0.988028069189867  \\
            3.33  0.9881039751676618  \\
            3.34  0.9881791468810934  \\
            3.35  0.988253593968042  \\
            3.36  0.9883273259054003  \\
            3.37  0.9884003520123591  \\
            3.38  0.9884726814536132  \\
            3.39  0.9885443232424902  \\
            3.4  0.9886152862440033  \\
            3.41  0.9886855791778316  \\
            3.42  0.9887552106212293  \\
            3.43  0.9888241890118656  \\
            3.44  0.9888925226505976  \\
            3.45  0.988960219704177  \\
            3.46  0.9890272882078952  \\
            3.47  0.9890937360681643  \\
            3.48  0.989159571065039  \\
            3.49  0.9892248008546806  \\
            3.5  0.9892894329717629  \\            
        };

\addplot [color=green!50!black,dashed,line width=1pt,forget plot]
  table[row sep=crcr]{%
            1.405  0.0  \\
           1.4051 0.8982380995141006 \\
};

\end{axis}
\end{tikzpicture}
    \begin{tikzpicture}

\begin{axis}[%
width=6cm,
height=2cm,
at={(0cm,0cm)},
scale only axis,
separate axis lines,
% every outer x axis line/.append style={black},
% every x tick label/.append style={font=\color{black}},
ymin=1.5,
ymax=3.65,
xlabel={$\lambda$},
ylabel={$\mu^\star_4(\lambda)$},
xmajorgrids,
% every outer y axis line/.append style={black},
% every y tick label/.append style={font=\color{black}},
xmin=0,
xmax=3.5,
extra x ticks={1.4050},
extra x tick labels={\; $\lambc$},
extra tick style={grid=major, grid style={dashed, gray}},
ymajorgrids,
title style={font=\small},
legend style={at={(0.97,0.03)},anchor=south east,legend cell align=left,align=left,draw=black},
legend columns = 2,
xticklabel shift={.1cm},
legend style={font=\tiny},
yticklabel shift={.0cm},
title={spectral norm, $d=4$}
]

\addplot+[no markers,color=red!70!black,line width=1,solid] 
        table[row sep={\\}]
        {
            x  y  \\
            0.0  1.7904299422486885  \\
            1.4  1.7904299422486885  \\
            1.41  1.7968964389737527  \\
            1.42  1.8034478248883317  \\
            1.43  1.810079398241571  \\
            1.44  1.8167869765934401  \\
            1.45  1.8235668060959658  \\
            1.46  1.8304154919100826  \\
            1.47  1.8373299437737818  \\
            1.48  1.8443073326927626  \\
            1.49  1.8513450559660038  \\
            1.5  1.8584407085710257  \\
            1.51  1.8655920594796347  \\
            1.52  1.8727970318507432  \\
            1.53  1.8800536863108819  \\
            1.54  1.8873602067220676  \\
            1.55  1.8947148879742977  \\
            1.56  1.9021161254416559  \\
            1.57  1.9095624058172345  \\
            1.58  1.9170522990999213  \\
            1.59  1.9245844515504926  \\
            1.6  1.932157579468911  \\
            1.61  1.939770463671695  \\
            1.62  1.9474219445696013  \\
            1.63  1.9551109177628216  \\
            1.64  1.9628363300846232  \\
            1.65  1.9705971760353707  \\
            1.66  1.978392494557937  \\
            1.67  1.986221366112879  \\
            1.68  1.994082910017865  \\
            1.69  2.001976282020923  \\
            1.7  2.0099006720812884  \\
            1.71  2.0178553023351933  \\
            1.72  2.0258394252269207  \\
            1.73  2.0338523217879825  \\
            1.74  2.041893300049425  \\
            1.75  2.0499616935741125  \\
            1.76  2.058056860097412  \\
            1.77  2.0661781802660477  \\
            1.78  2.0743250564660767  \\
            1.79  2.0824969117319503  \\
            1.8  2.0906931887294897  \\
            1.81  2.0989133488063945  \\
            1.82  2.1071568711045603  \\
            1.83  2.1154232517290796  \\
            1.84  2.1237120029693015  \\
            1.85  2.132022652567809  \\
            1.86  2.1403547430335546  \\
            1.87  2.148707830995754  \\
            1.88  2.157081486595472  \\
            1.89  2.1654752929120993  \\
            1.9  2.1738888454221765  \\
            1.91  2.1823217514882565  \\
            1.92  2.1907736298756824  \\
            1.93  2.1992441102953517  \\
            1.94  2.207732832970696  \\
            1.95  2.216239448227251  \\
            1.96  2.2247636161033313  \\
            1.97  2.233305005980428  \\
            1.98  2.2418632962320753  \\
            1.99  2.250438173890023  \\
            2.0  2.259029334326628  \\
            2.01  2.2676364809524845  \\
            2.02  2.276259324928364  \\
            2.03  2.2848975848906146  \\
            2.04  2.2935509866892345  \\
            2.05  2.302219263137877  \\
            2.06  2.3109021537751078  \\
            2.07  2.3195994046362816  \\
            2.08  2.328310768035446  \\
            2.09  2.3370360023567147  \\
            2.1  2.345774871854604  \\
            2.11  2.354527146462848  \\
            2.12  2.3632926016112386  \\
            2.13  2.3720710180500775  \\
            2.14  2.3808621816818407  \\
            2.15  2.3896658833996844  \\
            2.16  2.398481918932455  \\
            2.17  2.407310088695866  \\
            2.18  2.416150197649547  \\
            2.19  2.425002055159666  \\
            2.2  2.433865474866871  \\
            2.21  2.4427402745592754  \\
            2.22  2.451626276050269  \\
            2.23  2.4605233050609083  \\
            2.24  2.4694311911066933  \\
            2.25  2.4783497673885093  \\
            2.26  2.4872788706875593  \\
            2.27  2.496218341264099  \\
            2.28  2.5051680227598045  \\
            2.29  2.5141277621036195  \\
            2.3  2.523097409420916  \\
            2.31  2.532076817945843  \\
            2.32  2.5410658439367024  \\
            2.33  2.5500643465942487  \\
            2.34  2.5590721879827685  \\
            2.35  2.568089232953832  \\
            2.36  2.577115349072608  \\
            2.37  2.586150406546635  \\
            2.38  2.595194278156933  \\
            2.39  2.6042468391913975  \\
            2.4  2.6133079673803388  \\
            2.41  2.6223775428341214  \\
            2.42  2.631455447982792  \\
            2.43  2.6405415675176345  \\
            2.44  2.6496357883345674  \\
            2.45  2.6587379994793254  \\
            2.46  2.6678480920943355  \\
            2.47  2.676965959367253  \\
            2.48  2.6860914964810574  \\
            2.49  2.695224600565693  \\
            2.5  2.704365170651153  \\
            2.51  2.7135131076219943  \\
            2.52  2.722668314173199  \\
            2.53  2.7318306947673476  \\
            2.54  2.741000155593065  \\
            2.55  2.7501766045246696  \\
            2.56  2.759359951083011  \\
            2.57  2.768550106397431  \\
            2.58  2.7777469831688224  \\
            2.59  2.7869504956337416  \\
            2.6  2.796160559529543  \\
            2.61  2.805377092060491  \\
            2.62  2.814600011864827  \\
            2.63  2.823829238982755  \\
            2.64  2.833064694825312  \\
            2.65  2.8423063021440935  \\
            2.66  2.8515539850018157  \\
            2.67  2.860807668743667  \\
            2.68  2.870067279969449  \\
            2.69  2.8793327465064524  \\
            2.7  2.8886039973830724  \\
            2.71  2.897880962803117  \\
            2.72  2.9071635741207946  \\
            2.73  2.916451763816368  \\
            2.74  2.925745465472436  \\
            2.75  2.9350446137508364  \\
            2.76  2.944349144370147  \\
            2.77  2.953658994083766  \\
            2.78  2.9629741006585477  \\
            2.79  2.972294402853995  \\
            2.8  2.9816198404019625  \\
            2.81  2.990950353986886  \\
            2.82  3.0002858852264955  \\
            2.83  3.0096263766530233  \\
            2.84  3.0189717716948605  \\
            2.85  3.0283220146586816  \\
            2.86  3.037677050712001  \\
            2.87  3.0470368258661558  \\
            2.88  3.0564012869597064  \\
            2.89  3.065770381642233  \\
            2.9  3.075144058358528  \\
            2.91  3.0845222663331637  \\
            2.92  3.093904955555426  \\
            2.93  3.103292076764617  \\
            2.94  3.112683581435684  \\
            2.95  3.122079421765209  \\
            2.96  3.1314795506577084  \\
            2.97  3.1408839217122675  \\
            2.98  3.1502924892094626  \\
            2.99  3.1597052080986137  \\
            3.0  3.1691220339853037  \\
            3.01  3.1785429231191986  \\
            3.02  3.1879678323821397  \\
            3.03  3.197396719276502  \\
            3.04  3.206829541913828  \\
            3.05  3.216266259003698  \\
            3.06  3.2257068298428653  \\
            3.07  3.235151214304625  \\
            3.08  3.2445993728284215  \\
            3.09  3.2540512664096792  \\
            3.1  3.2635068565898733  \\
            3.11  3.2729661054467893  \\
            3.12  3.282428975585026  \\
            3.13  3.2918954301266816  \\
            3.14  3.3013654327022532  \\
            3.15  3.3108389474417232  \\
            3.16  3.3203159389658485  \\
            3.17  3.32979637237762  \\
            3.18  3.3392802132539203  \\
            3.19  3.3487674276373403  \\
            3.2  3.358257982028185  \\
            3.21  3.367751843376635  \\
            3.22  3.3772489790750764  \\
            3.23  3.386749356950592  \\
            3.24  3.3962529452576042  \\
            3.25  3.40575971267067  \\
            3.26  3.4152696282774295  \\
            3.27  3.4247826615716948  \\
            3.28  3.434298782446674  \\
            3.29  3.4438179611883495  \\
            3.3  3.453340168468971  \\
            3.31  3.4628653753406926  \\
            3.32  3.4723935532293333  \\
            3.33  3.4819246739282628  \\
            3.34  3.4914587095924103  \\
            3.35  3.500995632732388  \\
            3.36  3.510535416208739  \\
            3.37  3.5200780332262904  \\
            3.38  3.5296234573286176  \\
            3.39  3.539171662392629  \\
            3.4  3.548722622623239  \\
            3.41  3.5582763125481556  \\
            3.42  3.567832707012764  \\
            3.43  3.5773917811751175  \\
            3.44  3.5869535105010057  \\
            3.45  3.5965178707591434  \\
            3.46  3.606084838016423  \\
            3.47  3.615654388633285  \\
            3.48  3.62522649925915  \\
            3.49  3.63480114682796  \\
            3.5  3.644378308553784  \\
        };

\addplot+[no markers,dashed,domain=0:3.5,color=black,thick] {x};

 \node[] at (axis cs: 2.6,2.1) {$y = \lambda$};

\end{axis}
\end{tikzpicture}
    
    \begin{tikzpicture}

\begin{axis}[%
width=6cm,
height=2cm,
at={(0cm,0cm)},
scale only axis,
separate axis lines,
% every outer x axis line/.append style={black},
% every x tick label/.append style={font=\color{black}},
ymin=0,
ymax=1,
xlabel={$\lambda$},
ylabel={$|\scalar{x}{\xml(\lambda)}|$},
xmajorgrids,
% every outer y axis line/.append style={black},
% every y tick label/.append style={font=\color{black}},
xmin=0,
xmax=3.5,
extra x ticks={1.5320},
extra x tick labels={\; $\lambc$},
extra tick style={grid=major, grid style={dashed, gray}},
ytick={0, 0.2, 0.4, 0.6, 0.8, 1},
ymajorgrids,
title style={font=\small},
legend style={at={(0.97,0.03)},anchor=south east,legend cell align=left,align=left,draw=black},
legend columns = 2,
xticklabel shift={.1cm},
legend style={font=\tiny},
yticklabel shift={.0cm},
title={alignment, $d=5$}
]

\addplot [color=green!50!black,solid,line width=1.8pt,forget plot]
  table[row sep=crcr]{%
            0.0  0.0  \\
            1.5320  0.0  \\
};

\addplot+[no markers,color=green!50!black,line width=1,solid] 
        table[row sep={\\}]
        {
            x  y  \\
            1.54  0.9345820895706293  \\
            1.55  0.9360988162049152  \\
            1.56  0.9375427073548503  \\
            1.57  0.9389199436949902  \\
            1.58  0.9402359173705606  \\
            1.59  0.9414953657142194  \\
            1.6  0.9427024770996622  \\
            1.61  0.9438609757244542  \\
            1.62  0.9449741902388752  \\
            1.63  0.9460451098355542  \\
            1.64  0.9470764304951655  \\
            1.65  0.9480705934238502  \\
            1.66  0.9490298172380294  \\
            1.67  0.9499561250984004  \\
            1.68  0.9508513677308297  \\
            1.69  0.9517172430725946  \\
            1.7  0.9525553131305143  \\
            1.71  0.9533670185205719  \\
            1.72  0.9541536910678196  \\
            1.73  0.9549165647742366  \\
            1.74  0.9556567854060812  \\
            1.75  0.9563754189076442  \\
            1.76  0.957073458812594  \\
            1.77  0.9577518327953065  \\
            1.78  0.9584114084812447  \\
            1.79  0.9590529986164222  \\
            1.8  0.9596773656803792  \\
            1.81  0.9602852260142561  \\
            1.82  0.9608772535248895  \\
            1.83  0.9614540830170057  \\
            1.84  0.9620163131981814  \\
            1.85  0.9625645093950249  \\
            1.86  0.9630992060138008  \\
            1.87  0.9636209087742842  \\
            1.88  0.9641300967418768  \\
            1.89  0.9646272241798077  \\
            1.9  0.9651127222405048  \\
            1.91  0.9655870005128705  \\
            1.92  0.9660504484401755  \\
            1.93  0.9665034366215385  \\
            1.94  0.9669463180084503  \\
            1.95  0.9673794290064873  \\
            1.96  0.9678030904912249  \\
            1.97  0.9682176087463601  \\
            1.98  0.9686232763311868  \\
            1.99  0.9690203728838024  \\
            2.0  0.9694091658657553  \\
            2.01  0.9697899112532528  \\
            2.02  0.9701628541795262  \\
            2.03  0.9705282295324938  \\
            2.04  0.9708862625114479  \\
            2.05  0.9712371691461357  \\
            2.06  0.97158115678128  \\
            2.07  0.9719184245292944  \\
            2.08  0.9722491636936995  \\
            2.09  0.9725735581655081  \\
            2.1  0.9728917847946498  \\
            2.11  0.9732040137383173  \\
            2.12  0.9735104087879517  \\
            2.13  0.9738111276764351  \\
            2.14  0.9741063223669273  \\
            2.15  0.974396139324657  \\
            2.16  0.9746807197728729  \\
            2.17  0.9749601999340608  \\
            2.18  0.9752347112574401  \\
            2.19  0.9755043806336721  \\
            2.2  0.9757693305976444  \\
            2.21  0.9760296795201183  \\
            2.22  0.9762855417889742  \\
            2.23  0.9765370279807274  \\
            2.24  0.9767842450229408  \\
            2.25  0.9770272963481096  \\
            2.26  0.9772662820395542  \\
            2.27  0.9775012989698174  \\
            2.28  0.9777324409320243  \\
            2.29  0.9779597987646329  \\
            2.3  0.9781834604699733  \\
            2.31  0.978403511326942  \\
            2.32  0.9786200339981979  \\
            2.33  0.9788331086321773  \\
            2.34  0.9790428129602297  \\
            2.35  0.9792492223891481  \\
            2.36  0.9794524100893597  \\
            2.37  0.9796524470790128  \\
            2.38  0.9798494023041942  \\
            2.39  0.9800433427154839  \\
            2.4  0.9802343333410493  \\
            2.41  0.9804224373564625  \\
            2.42  0.980607716151419  \\
            2.43  0.9807902293935176  \\
            2.44  0.9809700350892581  \\
            2.45  0.9811471896424002  \\
            2.46  0.9813217479098189  \\
            2.47  0.9814937632549846  \\
            2.48  0.9816632875991883  \\
            2.49  0.981830371470623  \\
            2.5  0.9819950640514309  \\
            2.51  0.9821574132228128  \\
            2.52  0.9823174656082976  \\
            2.53  0.9824752666152593  \\
            2.54  0.9826308604747659  \\
            2.55  0.9827842902798402  \\
            2.56  0.9829355980222064  \\
            2.57  0.9830848246275959  \\
            2.58  0.9832320099896754  \\
            2.59  0.983377193002665  \\
            2.6  0.9835204115927044  \\
            2.61  0.9836617027480229  \\
            2.62  0.9838011025479694  \\
            2.63  0.9839386461909525  \\
            2.64  0.9840743680213383  \\
            2.65  0.9842083015553539  \\
            2.66  0.9843404795060372  \\
            2.67  0.9844709338072773  \\
            2.68  0.9845996956369834  \\
            2.69  0.9847267954394187  \\
            2.7  0.9848522629467376  \\
            2.71  0.9849761271997554  \\
            2.72  0.9850984165679882  \\
            2.73  0.9852191587689884  \\
            2.74  0.9853383808870074  \\
            2.75  0.985456109391014  \\
            2.76  0.9855723701520919  \\
            2.77  0.985687188460244  \\
            2.78  0.9858005890406264  \\
            2.79  0.9859125960692358  \\
            2.8  0.9860232331880693  \\
            2.81  0.9861325235197814  \\
            2.82  0.9862404896818552  \\
            2.83  0.9863471538003065  \\
            2.84  0.9864525375229418  \\
            2.85  0.986556662032184  \\
            2.86  0.9866595480574837  \\
            2.87  0.9867612158873327  \\
            2.88  0.9868616853808937  \\
            2.89  0.9869609759792604  \\
            2.9  0.987059106716363  \\
            2.91  0.9871560962295308  \\
            2.92  0.9872519627697256  \\
            2.93  0.9873467242114583  \\
            2.94  0.9874403980623975  \\
            2.95  0.9875330014726864  \\
            2.96  0.987624551243973  \\
            2.97  0.987715063838167  \\
            2.98  0.9878045553859333  \\
            2.99  0.9878930416949291  \\
            3.0  0.9879805382577971  \\
            3.01  0.9880670602599191  \\
            3.02  0.9881526225869426  \\
            3.03  0.9882372398320852  \\
            3.04  0.9883209263032259  \\
            3.05  0.9884036960297896  \\
            3.06  0.9884855627694342  \\
            3.07  0.9885665400145431  \\
            3.08  0.9886466409985332  \\
            3.09  0.9887258787019831  \\
            3.1  0.9888042658585876  \\
            3.11  0.9888818149609434  \\
            3.12  0.9889585382661743  \\
            3.13  0.9890344478013974  \\
            3.14  0.9891095553690388  \\
            3.15  0.9891838725520017  \\
            3.16  0.9892574107186927  \\
            3.17  0.9893301810279099  \\
            3.18  0.9894021944335988  \\
            3.19  0.9894734616894777  \\
            3.2  0.98954399335354  \\
            3.21  0.9896137997924341  \\
            3.22  0.9896828911857273  \\
            3.23  0.9897512775300552  \\
            3.24  0.9898189686431622  \\
            3.25  0.9898859741678343  \\
            3.26  0.98995230357573  \\
            3.27  0.9900179661711098  \\
            3.28  0.9900829710944697  \\
            3.29  0.9901473273260799  \\
            3.3  0.9902110436894324  \\
            3.31  0.9902741288546008  \\
            3.32  0.9903365913415132  \\
            3.33  0.9903984395231424  \\
            3.34  0.9904596816286158  \\
            3.35  0.9905203257462449  \\
            3.36  0.9905803798264818  \\
            3.37  0.9906398516847991  \\
            3.38  0.9906987490044988  \\
            3.39  0.9907570793394531  \\
            3.4  0.9908148501167743  \\
            3.41  0.9908720686394232  \\
            3.42  0.9909287420887498  \\
            3.43  0.9909848775269748  \\
            3.44  0.991040481899609  \\
            3.45  0.9910955620378155  \\
            3.46  0.9911501246607143  \\
            3.47  0.9912041763776317  \\
            3.48  0.9912577236902965  \\
            3.49  0.9913107729949834  \\
            3.5  0.9913633305846051  \\           
        };

\addplot [color=green!50!black,dashed,line width=1pt,forget plot]
  table[row sep=crcr]{%
            1.5320  0.0  \\
           1.5321 0.9345820895706293 \\
};

\end{axis}
\end{tikzpicture}
    \begin{tikzpicture}

\begin{axis}[%
width=6cm,
height=2cm,
at={(0cm,0cm)},
scale only axis,
separate axis lines,
% every outer x axis line/.append style={black},
% every x tick label/.append style={font=\color{black}},
ymin=1.5,
ymax=3.65,
xlabel={$\lambda$},
ylabel={$\mu^\star_5(\lambda)$},
xmajorgrids,
% every outer y axis line/.append style={black},
% every y tick label/.append style={font=\color{black}},
xmin=0,
xmax=3.5,
extra x ticks={1.5320},
extra x tick labels={\; $\lambc$},
extra tick style={grid=major, grid style={dashed, gray}},
ymajorgrids,
title style={font=\small},
legend style={at={(0.97,0.03)},anchor=south east,legend cell align=left,align=left,draw=black},
legend columns = 2,
xticklabel shift={.1cm},
legend style={font=\tiny},
yticklabel shift={.0cm},
title={spectral norm, $d=5$}
]

\addplot [color=red!70!black,solid,line width=1,forget plot]
  table[row sep=crcr]{%
            0.0  1.886390140535112  \\
            1.54  1.886390140535112  \\
};

\addplot+[no markers,domain=1.207:3.5,color=red!70!black,line width=1,solid,samples=100] 
        table[row sep={\\}]
        {
            x  y  \\
            1.54  1.8934899852261786  \\
            1.55  1.9006491817525923  \\
            1.56  1.907865186924751  \\
            1.57  1.915135661083122  \\
            1.58  1.922458442443433  \\
            1.59  1.9298315258336232  \\
            1.6  1.9372530448916332  \\
            1.61  1.9447212570245174  \\
            1.62  1.9522345305956046  \\
            1.63  1.9597913339277708  \\
            1.64  1.9673902258008102  \\
            1.65  1.97502984718846  \\
            1.66  1.9827089140319782  \\
            1.67  1.9904262108866844  \\
            1.68  1.9981805853085948  \\
            1.69  2.00597094287234  \\
            1.7  2.0137962427306806  \\
            1.71  2.0216554936411053  \\
            1.72  2.0295477503973065  \\
            1.73  2.037472110613213  \\
            1.74  2.0454277118153854  \\
            1.75  2.053413728806226  \\
            1.76  2.0614293712659437  \\
            1.77  2.069473881565782  \\
            1.78  2.0775465327688307  \\
            1.79  2.0856466267979425  \\
            1.8  2.093773492752954  \\
            1.81  2.1019264853617377  \\
            1.82  2.1101049835514942  \\
            1.83  2.118308389128417  \\
            1.84  2.126536125555236  \\
            1.85  2.1347876368174026  \\
            1.86  2.143062386369719  \\
            1.87  2.1513598561561422  \\
            1.88  2.1596795456962865  \\
            1.89  2.168020971232834  \\
            1.9  2.176383664934692  \\
            1.91  2.184767174151253  \\
            1.92  2.1931710607135724  \\
            1.93  2.20159490027873  \\
            1.94  2.2100382817139637  \\
            1.95  2.2185008065175116  \\
            1.96  2.2269820882733855  \\
            1.97  2.2354817521375447  \\
            1.98  2.243999434353178  \\
            1.99  2.2525347817930017  \\
            2.0  2.2610874515266577  \\
            2.01  2.2696571104114707  \\
            2.02  2.278243434704974  \\
            2.03  2.286846109697715  \\
            2.04  2.2954648293650335  \\
            2.05  2.304099296036525  \\
            2.06  2.3127492200821083  \\
            2.07  2.321414319613593  \\
            2.08  2.3300943202008235  \\
            2.09  2.3387889546014655  \\
            2.1  2.347497962503639  \\
            2.11  2.356221090280604  \\
            2.12  2.3649580907568057  \\
            2.13  2.3737087229846123  \\
            2.14  2.3824727520311244  \\
            2.15  2.391249948774496  \\
            2.16  2.4000400897092256  \\
            2.17  2.4088429567599277  \\
            2.18  2.4176583371031244  \\
            2.19  2.426486022996609  \\
            2.2  2.4353258116160044  \\
            2.21  2.444177504898107  \\
            2.22  2.4530409093906953  \\
            2.23  2.46191583610844  \\
            2.24  2.4708021003946357  \\
            2.25  2.4796995217884295  \\
            2.26  2.488607923897305  \\
            2.27  2.4975271342745398  \\
            2.28  2.5064569843014093  \\
            2.29  2.5153973090738933  \\
            2.3  2.5243479472936947  \\
            2.31  2.5333087411633435  \\
            2.32  2.542279536285203  \\
            2.33  2.5512601815642086  \\
            2.34  2.560250529114154  \\
            2.35  2.569250434167376  \\
            2.36  2.5782597549876733  \\
            2.37  2.587278352786329  \\
            2.38  2.596306091641091  \\
            2.39  2.6053428384179846  \\
            2.4  2.614388462695839  \\
            2.41  2.623442836693402  \\
            2.42  2.63250583519894  \\
            2.43  2.641577335502227  \\
            2.44  2.6506572173288  \\
            2.45  2.659745362776412  \\
            2.46  2.66884165625358  \\
            2.47  2.6779459844201416  \\
            2.48  2.687058236129747  \\
            2.49  2.6961783023742045  \\
            2.5  2.705306076229606  \\
            2.51  2.7144414528041665  \\
            2.52  2.723584329187702  \\
            2.53  2.7327346044026934  \\
            2.54  2.7418921793568742  \\
            2.55  2.7510569567972727  \\
            2.56  2.760228841265665  \\
            2.57  2.7694077390553886  \\
            2.58  2.77859355816945  \\
            2.59  2.7877862082798974  \\
            2.6  2.7969856006884046  \\
            2.61  2.8061916482880074  \\
            2.62  2.8154042655259883  \\
            2.63  2.8246233683678197  \\
            2.64  2.8338488742621744  \\
            2.65  2.843080702106928  \\
            2.66  2.852318772216149  \\
            2.67  2.861563006288018  \\
            2.68  2.8708133273736554  \\
            2.69  2.880069659846831  \\
            2.7  2.8893319293745168  \\
            2.71  2.8986000628882467  \\
            2.72  2.907873988556289  \\
            2.73  2.9171536357565557  \\
            2.74  2.9264389350502715  \\
            2.75  2.935729818156342  \\
            2.76  2.945026217926423  \\
            2.77  2.954328068320647  \\
            2.78  2.9636353043840016  \\
            2.79  2.9729478622233394  \\
            2.8  2.982265678984979  \\
            2.81  2.9915886928328983  \\
            2.82  3.000916842927506  \\
            2.83  3.0102500694049446  \\
            2.84  3.0195883133569357  \\
            2.85  3.028931516811146  \\
            2.86  3.0382796227120408  \\
            2.87  3.047632574902228  \\
            2.88  3.056990318104284  \\
            2.89  3.0663527979030087  \\
            2.9  3.0757199607281502  \\
            2.91  3.085091753837538  \\
            2.92  3.094468125300642  \\
            2.93  3.1038490239825376  \\
            2.94  3.1132343995282454  \\
            2.95  3.1226242023474775  \\
            2.96  3.1320183835997297  \\
            2.97  3.1414168951797397  \\
            2.98  3.150819689703298  \\
            2.99  3.1602267204933927  \\
            3.0  3.16963794156668  \\
            3.01  3.179053307620281  \\
            3.02  3.188472774018877  \\
            3.03  3.1978962967821154  \\
            3.04  3.207323832572314  \\
            3.05  3.2167553386824244  \\
            3.06  3.226190773024306  \\
            3.07  3.2356300941172433  \\
            3.08  3.245073261076735  \\
            3.09  3.2545202336035373  \\
            3.1  3.263970971972959  \\
            3.11  3.273425437024381  \\
            3.12  3.2828835901510356  \\
            3.13  3.2923453932899878  \\
            3.14  3.301810808912361  \\
            3.15  3.311279800013759  \\
            3.16  3.320752330104923  \\
            3.17  3.330228363202561  \\
            3.18  3.3397078638204083  \\
            3.19  3.3491907969604666  \\
            3.2  3.358677128104427  \\
            3.21  3.3681668232052937  \\
            3.22  3.3776598486791736  \\
            3.23  3.387156171397251  \\
            3.24  3.396655758677925  \\
            3.25  3.406158578279114  \\
            3.26  3.41566459839073  \\
            3.27  3.4251737876273034  \\
            3.28  3.434686115020763  \\
            3.29  3.4442015500133705  \\
            3.3  3.4537200624507918  \\
            3.31  3.46324162257533  \\
            3.32  3.4727662010192732  \\
            3.33  3.482293768798397  \\
            3.34  3.4918242973055924  \\
            3.35  3.5013577583046254  \\
            3.36  3.5108941239240172  \\
            3.37  3.5204333666510634  \\
            3.38  3.5299754593259474  \\
            3.39  3.5395203751359965  \\
            3.4  3.5490680876100344  \\
            3.41  3.558618570612869  \\
            3.42  3.5681717983398515  \\
            3.43  3.5777277453115848  \\
            3.44  3.5872863863687035  \\
            3.45  3.596847696666775  \\
            3.46  3.6064116516712903  \\
            3.47  3.6159782271527585  \\
            3.48  3.625547399181885  \\
            3.49  3.6351191441248587  \\
            3.5  3.6446934386387158  \\
        }
        ;

\addplot+[no markers,dashed,domain=0:3.5,color=black,thick] {x};

 \node[] at (axis cs: 2.6,2.1) {$y = \lambda$};

\end{axis}
\end{tikzpicture}    
    
    \caption{Phase transition for the ML estimator of the symmetric rank-one spiked tensor model \eqref{SRO} with $d=3,4,5$, according to \cite{JagaLM-20-AAP}: (a) asymptotic alignment between $x$ and ML solution $\xml(\lambda)$; (b) asymptotic maximum likelihood (or spectral norm of $\tensor{Y}$). The argument of $\lambc(d)$ has been omitted for simplicity.}
    \label{fig:align-spnorm}
\end{figure}

For $d=3,4,5$, one can analytically solve $\frac{\partial \phi_d(\lambda, t)}{\partial t} = d \lambda^2 t^{d-1} + t/(t-1) = 0$ to find $q_d(\lambda)$ for $\lambc(d)$, which gives
\begin{align}
\label{form-q3}
 q_3(\lambda) = & \ \frac{1}{2} + \sqrt{\frac{3 \lambda^2 - 4}{12 \lambda^2}},
 \quad \lambda > \lambc(3), \\
 q_4(\lambda) = & \ 
  \frac{1}{3} 
   + \frac{g_4(\lambda)}{6} 
   + \frac{2}{3 g_4(\lambda)}, 
 \quad \lambda > \lambc(4), \\
\label{form-q4} 
 q_5(\lambda) = & \ \frac{1}{4}+\frac{\sqrt{15}\, h_5(\lambda)}{60}+\frac{\sqrt{450}}{60} \sqrt{\frac{\left(\lambda  g_5(\lambda)^{\frac{1}{3}}-\frac{g_5(\lambda)^{\frac{2}{3}}}{15}-16\right) h_5(\lambda)+\sqrt{15}\, \lambda  g_5(\lambda)^{\frac{1}{3}}}{g_5(\lambda)^{\frac{1}{3}} h_5(\lambda)\, \lambda}}, \quad \lambda > \lambc(5),
 \label{form-q5} 
\end{align}
with
\begin{align}
 g_4(\lambda) = & \ \left( \frac{8\lambda^2 + 3\sqrt{81 - 48\lambda^2} - 27 }{\lambda^2} \right)^\frac{1}{3}, \\
 g_5(\lambda) = & \ 60 \sqrt{15}\, \sqrt{135 \lambda^{2}-256}+2700 \lambda, \\
 h_5(\lambda) = & \ \sqrt{\frac{2 g_5(\lambda)^{\frac{2}{3}}+15 \lambda  g_5(\lambda)^{\frac{1}{3}}+480}{\lambda  g_5(\lambda)^{\frac{1}{3}}}}, 
\end{align}
and hence
\begin{align}
 \mu^\star_3(\lambda) = & \ \frac{3\lambda^2 + \lambda\sqrt{9\lambda^2 - 12}+4}{\sqrt{18\lambda^2 + 6\lambda\sqrt{9\lambda^2 - 12}}},
 \quad \lambda > \lambc(3), \\
 \mu^\star_4(\lambda) = & \ 
  \frac{2+2\lambda^{2} \left(\frac{1}{3} 
   + \frac{h_4(\lambda)}{6} 
   + \frac{2}{3h_4(\lambda)}\right)^{3}}{\sqrt{1+4 \lambda^{2} \left(\frac{1}{3} + \frac{h_4(\lambda)}{6} 
   + \frac{2}{3h_4(\lambda)}\right)^{3}}}, 
 \quad \lambda > \lambc(4).
\end{align}
These quantities are plotted in Figure \ref{fig:align-spnorm}.
(As the expression of $\mu^\star_5$ is too long, we choose to omit it here.)
In fact, one can easily check that the above explicit formulas are still algebraically valid slightly below $\lambc(d)$, namely, for $\lambs(d) \le \lambda < \lambc(d)$ with $\lambs(d) := \sqrt{(d-1)^{d-1}/d(d-2)^{d-2}}$. 
As discussed by \cite{JagaLM-20-AAP} and \cite{ArouMMN-19-CPAM}, a local maximizer of the likelihood having alignment given (asymptotically) by $\sqrt{q_d(\lambda)}$ with $x$ exists in this regime as $N$ grows large.
However, the above results are relevant not only to ML estimation, but in fact to any estimator: it turns out that the ML estimator is information-theoretically optimal, in the sense that no other estimator can attain a higher correlation $|\scalar{x}{\xml(\lambda)}|$ for any $\lambda \ge 0$, as also shown by \cite{JagaLM-20-AAP}. This implies, in particular, that no estimator $\hat{x}(\tensor{Y})$ of $x$ can (asymptotically) achieve nonzero alignment with $x$ in the regime $\lambda < \lambc(d)$ without additional information on $x$.
Therefore, the above mentioned local maximizer that is positively correlated with $x$ for $\lambda < \lambc(d)$ cannot be a valid estimator of $x$---that is, it cannot be expressed a measurable function of $\tensor{Y}$.

The reader will notice that the above discussion makes no reference to the spectral viewpoint discussed in the previous subsection. 
The omission is intentional, as the statement and derivation of the results of \citep{JagaLM-20-AAP} do not rely on or mention in any way the connection between ML estimation and tensor eigenpairs. 
By bringing this connection into the picture and combining it with our proposed approach described in the previous section, we are able to reach some of the same predictions but using standard tools from RMT rather than from spin glass theory.

% ================================================================

\subsection{Analysis under the light of random matrix theory}

In this section, we explain how we leverage our approach to obtain the result described by Theorem \ref{thm:main}, concerning ML estimation of the planted vector $x$ of the symmetric spiked rank-one tensor model \eqref{SRO}.
Then, we will discuss connections with other works, limitations and possible extensions of this result.
The proof of Theorem \ref{thm:main} itself is postponed to Section~\ref{sec:calc}.

\subsubsection{Analyzing contractions of critical points}
\label{sec:main}

Let us now turn back to the goal of studying $\tensor{Y} \cdot (\xml(\lambda)^{d-2})$. 
As there seems to be no simple way of taking into account the fact that $\xml(\lambda)$ is a global maximum, we consider more generally random contractions $\tensor{Y} \cdot u^{d-2}$ with critical points $u$.
This can be done by relying on the tensor eigenvalue equations as a characterization of critical points.
Nevertheless, these critical points must asymptotically behave as local maxima, since we also impose the condition that the largest eigenvalue of $\tensor{Y} \cdot u^{d-2}$ must (asymptotically) be strictly greater than $(d-1)\beta_d$ (cf.~Remark \ref{rem:mu}). The tensor eigenvalue equations give
\begin{equation}
 \mu \, u = \tensor{Y} \cdot u^{d-1}
           = \lambda \, \scalar{x}{u}^{d-1} \, x
            + \frac{1}{\sqrt{N}} \tensor{W} \cdot u^{d-1},
\end{equation} 
from which follows by taking scalar products with $u$ and $x$ (and recalling that $\norm{u}=\norm{x}=1$) that 
\begin{align}
\label{mu-start}
  \mu = & \ \lambda \, \scalar{x}{u}^{d}
            + \frac{1}{\sqrt{N}} \tensor{W} \cdot u^{d}, \\
  \scalar{x}{u} 
           = & \ \frac{\lambda}{\mu} \, \scalar{x}{u}^{d-1} 
            + \frac{1}{\mu \, \sqrt{N}} \scalar{x}{\tensor{W} \cdot u^{d-1}},
\label{align-start}            
\end{align}
respectively.
Taking the expectation of these expressions involves considerable difficulty, as $u$ depends on $\tensor{W}$ in an \emph{a priori} intricate manner, and thus cannot be pulled out from expectations.
The technical tools that allow managing this dependency are the implicit function theorem and Stein's lemma (also known simply as Gaussian integration by parts; see Appendix \ref{app:RMT-tools} for a statement). The former gives us the derivatives of $u$ and $\mu$ with respect to components of $\tensor{W}$, which read 
\begin{equation}
\label{deriv}
  \begin{pmatrix}
    \displaystyle \frac{\partial u}{\partial W_{i_1 \dots i_d}} \\[4mm]
    \displaystyle \frac{\partial \mu}{\partial W_{i_1 \dots i_d}}
  \end{pmatrix}
  = 
  \begin{pmatrix}
  \displaystyle
  - \frac{1}{(d-1)\sqrt{N}} R\left( \frac{\mu}{d-1} \right) \, \phi + \frac{1}{(d-2) \, \mu} \, \frac{\partial \mu}{\partial W_{i_1 \dots i_d}} \, u
 \\[4mm]
   \displaystyle \frac{1}{\sigma^2_{W_{i_1 \dots i_d}} \, \sqrt{N}} \, \prod_{j=1}^d u_{i_j}
  \end{pmatrix},
\end{equation} 
where $R(z)$ is the resolvent of $\tensor{Y} \cdot u^{d-2}$ (a definition is given in Appendix \ref{app:RMT-tools}) and $\phi$ is an $N$-dimensional vector that satisfies 
\begin{equation}
 \sigma^2_{W_{i_1 \dots i_d}} \, \phi
  = \frac{1}{d} \, \sum_{j=1}^d u_1 \dots u_{i_{j-1}} \, u_{i_{j+1}} \dots u_d \; e^{(i_j)},
\end{equation} 
with $e^{(i_j)}$ denoting the $i_j$ canonical basis vector.
The expectations on the right-hand sides of \eqref{mu-start} and \eqref{align-start} can then be computed by resorting to \eqref{deriv}, thanks to Stein's lemma.

Now, one must be careful when taking the large-$N$ limit of these expectations, because of the term involving the resolvent $R(z)$ evaluated at $z = \mu/(d-1)$ in \eqref{deriv}.
In fact, along the proof we show that 
\begin{equation}
   \forall \, z \in \mathbb{C} \text{ such that } \Im\{z\} > 0, 
    \quad  \frac{1}{N} \, \tr R(z) \convas m_d(z),
\end{equation} 
where $m_d(z)$ is as given by Theorem \ref{thm:spec-meas}, and thus by the Stieltjes continuity theorem (see Appendix \ref{app:RMT-tools}) the spectral measure of $\tensor{Y} \cdot u^{d-2}$ converges weakly almost surely to the same semicircle law of \eqref{sclaw-W}, which is supported on $[-\beta_d,\beta_d]$. 
Consequently, the sequence of critical points $u$ considered in Theorem \ref{thm:main} must be such that the almost sure limit $\mu_{d,\infty}(\lambda)$ of $\mu = \tensor{Y} \cdot u^{d}$ (assumed to exist) must be strictly greater that $(d-1)\beta_d$, so that $(d-1)^{-1} \, \mu_{d,\infty}(\lambda)$ does not fall on the limiting spectrum of $\tensor{Y} \cdot u^{d-2}$.

Once in possession of the asymptotic expressions of these expectations as $N \rightarrow \infty$, we invoke the assumption that the random quantities $\mu$ and $\scalar{x}{u}$ converge almost surely to a deterministic expression depending only on $\lambda$ and on $d$ (as assumed in the statement of Theorem \ref{thm:main}). 
This step (along with the fact that the normalized trace of the resolvent of $\tensor{Y} \cdot u^{d-2}$ also converges almost surely, as in Theorem \ref{thm:spec-meas}) allows breaking the resulting expectations of products into products of expectations, leading to the final fixed-point equation described in the statement of Theorem \ref{thm:main}. 
Moreover, it turns out that this equation effectively describes the performance of ML in the regime $\lambda > \lambc(d)$ for $d=3,4,5$, as we already mentioned and will further discuss next.

\subsubsection{Discussion}
\label{sec:disc}

One can readily verify that the expressions of $q_3$ and $\mu^\star_3$ given in Section \ref{sec:phasetran} precisely match those given by Lemma \ref{lem:fpeq}, meaning that our fixed point equation describes the asymptotic performance of the ML solution in the regime $\lambda > \lambc(3)$. Even though we did not compute the solution of our fixed-point equation explicitly for other values of $d$, it can be numerically verified for $d=4,5$ that they are again consistent with the characterization of \cite{JagaLM-20-AAP} described above. We therefore conjecture the following:

\begin{conjec}
 For any $d \ge 3$ and $\lambda > \lambc(d)$, the quantities characterized by Theorem \ref{thm:main} match those described in Section \ref{sec:phasetran}, that is,
 \begin{equation}
 \label{conj-eqs}
  \alpha_{d,\infty}(\lambda) = \sqrt{q_d(\lambda)}, \qquad
   \mu_{d,\infty}(\lambda) = \mu^\star_d(\lambda).
  \label{conj-mu} 
 \end{equation} 
 Furthermore, the first above equation also holds in the regime $\lambs(d) \le \lambda \le \lambc(d)$.
\end{conjec}

It can seem somewhat surprising that our analysis is able to predict the performance of ML estimation in the regime $\lambda > \lambc(d)$, despite relying on the tensor eigenvalue equations satisfied not only by the ML solution but also by all other critical points of the ML problem---which are on average exponentially numerous, according to the results of \cite{ArouMMN-19-CPAM}.
The key to this ``selectivity'' lies of course in the additional assumptions of Theorem \ref{thm:main}, which seem to restrict the considered sequence of critical points $u$ to asymptotically behave as global maxima of the ML problem in the regime $\lambda > \lambc(d)$.
However, a precise understanding of this effect remains yet to be achieved.

Another problem that remains open is whether our RMT-based approach can also predict the phase transition point $\lambc(d)$.
This being said, from a practical standpoint, it is also known that the ML solution cannot be numerically estimated in polynomial time for values of $\lambda$ close to $\lambc(d)$ (it is even conjectured that $\lambda$ must scale like $O(N^\ell)$ with $\ell > 0$ for polynomial-time algorithms to recover the solution \citep{ JagaLM-20-AAP, HopkSS-15-JMLR}); this, in a way, makes the very notion of a phase transition for the ML solution unessential.

%%%%%%%%%%%%%%%%%%%%%%%%%%%%%%%%%%%%%%%%%%%%%%%%%%%%%%%%%%%%%%%
%%%%%%%%%%%%%%%%%%%%%%%%%%%%%%%%%%%%%%%%%%%%%%%%%%%%%%%%%%%%%%%
%%%%%%%%%%%%%%%%%%%%%%%%%%%%%%%%%%%%%%%%%%%%%%%%%%%%%%%%%%%%%%%

\section{Proof of the main results}
\label{sec:calc}

In this section, we will provide the proof of Theorem \ref{thm:main}, and explain how one part of it can be adapted to show also Theorem \ref{thm:spec-meas}. For clarity, we have organized the proof into separate subsections which deal with each required ingredient. 

Recall that we consider a sequence of critical points $(\mu, u)$ of the ML problem, and hence satisfy the tensor eigenvalue equations $\tensor{Y} \cdot u^{d-2} = \mu \, u$ and $\norm{u}=1$.
Let us start by computing the expectation of $\mu$ as given by \eqref{mu-start}.

\paragraph{Notational conventions:} Since we manipulate explicit summations over indices $i_1,\ldots,i_d$ arising from the contraction of $\tensor{Y}$ with $u$ and $x$, it is convenient to introduce some notation for avoiding too lengthy expressions.
Namely, we will abbreviate the indices $i_1,\ldots,i_d$ by a boldface notation $\boldsymbol{i}$ (and similarly for $\boldsymbol{j}$ and $\boldsymbol{\ell}$). Under this convention, we have for instance $W_{i_1,\ldots,i_d} = W_{\boldsymbol{i}}$ and $\sum_{i_1,\ldots,i_d} = \sum_{\boldsymbol{i}}$, where each $i_j$ runs from $1$ to $N$.
We will also simplify the symbol used for the variance of $W_{\boldsymbol{i}}$, writing it more compactly as $\sigma^2_{\boldsymbol{i}}$ instead of $\sigma^2_{W_{\boldsymbol{i}}}$.
Finally, the shorthand $\prod_{k \neq j}^{d} u_{i_k}$ will be used to denote $\prod_{k \in \{1,\ldots,d\}\setminus\{j\}} u_{i_k}$.
Similarly, $\prod_{k \not\in S}^{d} u_{i_k}$ stands for $\prod_{k \in \{1,\ldots,d\}\setminus S} u_{i_k}$.

\subsection{Expectation of $\mu$}

The starting point is the expression
\begin{equation}
  \E{\mu} = \lambda \, \E{ \scalar{x}{u}^{d} }
            + \frac{1}{\sqrt{N}} \E{ \tensor{W} \cdot u^{d}}.
\end{equation}
To develop it, we apply Stein's lemma (which is simply Gaussian integration by parts, see Appendix \ref{app:RMT-tools}) to the second term on the right-hand side, by relying on the derivative of $u$ with respect to components of $\tensor{W}$, as computed in Appendix \ref{app:deriv-eig}.
This step requires the following result, which may not be immediately clear. Its proof will be postponed to Appendix \ref{app:cont-diff}.

\begin{lem}
\label{lem:cont-diff}
 There exists an almost everywhere continuously differentiable function $G : \mathcal{S}^d(N) \rightarrow \mathbb{R}^{N+1}$ such that $G(\tensor{W}) = (\mu(\tensor{W}), u(\tensor{W}))$ is an eigenpair of $\tensor{W}$ (for almost every $\tensor{W}$).
\end{lem}

The components of $u(\tensor{W})$ are bounded and $\mu(\tensor{W})$ has polynomial growth, and hence Stein's lemma applies. We also note that all subsequent applications of this tool in the following sections can be justified in the same manner.

Now, according to our computation in Appendix \ref{app:deriv-eig}, the partial derivative $\frac{\partial u_{i_j}}{\partial W_{\boldsymbol{i}}}$ of a component of such a function $u$ is expressed in terms of the resolvent of $\tensor{Y} \cdot u^{d-2}$, that is, of $R(z) = \left(\tensor{Y} \cdot u^{d-2} - z I \right)^{-1}$, and also of a certain vector $\phi$ defined in \eqref{phi-vec}, which depends on $u$ and on the variance of $W_{\boldsymbol{i}}$.
For simplicity, we will use the notation $\bar{R} := R(\mu/(d-1))$, as in Appendix \ref{app:deriv-eig}.

With these definitions, we can write
\begin{align}
  \E{ \tensor{W} \cdot u^{d}}
  = & \ \sum_{\boldsymbol{i}} \E{ W_{\boldsymbol{i}} \, u_{i_1} \dots u_{i_d}} \\
  = & \  \sum_{\boldsymbol{i}} \sigma^2_{\boldsymbol{i}} \, 
        \sum_{j=1}^d \, \E{ \frac{\partial u_{i_j}}{\partial W_{\boldsymbol{i}}} 
                          \prod_{k \neq j}^d u_{i_k}}  \\
  = & \  \sum_{\boldsymbol{i}} \sigma^2_{\boldsymbol{i}} \, \sum_{j=1}^d
         \, \E{ -\frac{1}{(d-1) \, \sqrt{N}} [\bar{R} \phi]_{i_j} \prod_{k \neq j}^d u_{i_k}
         + \frac{1}{\sigma^2_{\boldsymbol{i}}} \frac{1}{(d-2)\mu\sqrt{N}} \prod_{k=1}^d u_{i_k}^2 }.
\end{align}
Using now \eqref{phi-vec} and $\norm{u}=1$, we get 
\begin{align}
  \E{ \tensor{W} \cdot u^{d}}
  = & \ 
  -\frac{1}{(d-1) \, \sqrt{N}} \sum_{\boldsymbol{i}}
    \sum_{j=1}^d \E{ \left( \prod_{k \neq j}^d u_{i_k} \right)
            \frac{1}{d} \sum_{\ell=1}^d \bar{R}_{i_j,i_\ell} \prod_{m \neq \ell}^d u_{i_m}}
    + \frac{d}{(d-2)\sqrt{N}} \E{\frac{1}{\mu}}.
\end{align}      
Now, for each $j$, the summation in $\ell$ in the first expectation contains exactly one term of the form $\bar{R}_{i_j,i_j} \prod_{k \neq j}^d u_{i_j}^2$, which gives rise to $\tr \bar{R}$ when summing over all indices $i_1,\ldots,i_d$. All the other $d-1$ terms are of the form $(\prod_{k \not\in \{j,\ell\}}^d u_{i_k}^2) u_{i_j} \, \bar{R}_{i_j,i_\ell} u_{i_\ell}$. Because $(1/(\mu - \mu/(d-1)),u)$ is an eigenpair of $\bar{R}$, summing each such term over the indices $i_1,\ldots,i_d$ yields 
\begin{equation}
 \sum_{\boldsymbol{i}} \left(\prod_{k \not\in \{j,\ell\}}^d u_{i_k}^2 \right) u_{i_j} \, \bar{R}_{i_j,i_\ell} u_{i_\ell}
 = \scalar{u}{\bar{R}u}
 = \frac{d-1}{(d-2)\mu}.
\end{equation} 
Hence, using these results we arrive at
\begin{align}      
\E{ \tensor{W} \cdot u^{d}}
  = & \  - \frac{1}{d (d-1) \, \sqrt{N}} \, \E{ d \tr \bar{R} + \frac{d(d-1)^2}{(d-2)\mu}}
    + \frac{d}{(d-2)\sqrt{N}} \E{\frac{1}{\mu}}  \\
  = & \ \frac{1}{(d-2)\sqrt{N}} \, \E{\frac{1}{\mu}} 
      - \frac{1}{(d-1)\sqrt{N}} \, \E{\tr \bar{R}}.
\end{align}
Therefore, we have shown that 
\begin{equation}
\label{exp-mu}
  \E{\mu} = \lambda \, \E{ \scalar{x}{u}^{d} }
            + \frac{1}{(d-2) N} \, \E{\frac{1}{\mu}} 
            - \frac{1}{(d-1)N}  \, \E{\tr \bar{R}}.
\end{equation}

\subsection{Expectation of $\scalar{x}{u}$}

Now let us carry out a similar computation for the expected alignment:
\begin{align}
  \E{ \scalar{x}{u} }
           = & \ \lambda \, \E{ \frac{1}{\mu} \, \scalar{x}{u}^{d-1} }
            + \frac{1}{\sqrt{N}} \E{ \frac{1}{\mu} \, \scalar{x}{\tensor{W} \cdot u^{d-1}}}.
\end{align}
By explicitly developing the scalar product of the last term, we write 
\begin{equation}
 \E{ \frac{1}{\mu} \, \scalar{x}{\tensor{W} \cdot u^{d-1}}}
 = \sum_{\boldsymbol{i}}
  \E{ \frac{1}{\mu} \, W_{\boldsymbol{i}} \, u_{i_1} \dots u_{i_{d-1}} x_{i_d} }.
\end{equation} 
Treating now $\mu$ and each component of $u$ as functions of the Gaussian tensor $\tensor{W}$, by Stein's lemma we obtain
\begin{align}
 \E{ \frac{1}{\mu} \, \scalar{x}{\tensor{W} \cdot u^{d-1}}} 
   = & \ \sum_{\boldsymbol{i}} \sigma^2_{\boldsymbol{i}} \, 
       \E{ - \frac{1}{\mu^2} \, \frac{\partial \mu}{\partial W_{\boldsymbol{i}}}
          \, \left( \prod_{k=1}^{d-1} u_{i_k} \right) x_{i_d}
    + \frac{1}{\mu} x_{i_d} \sum_{j=1}^{d-1} \frac{\partial u_{i_j}}{\partial W_{\boldsymbol{i}}} 
      \prod_{k \neq j}^{d-1} u_{i_k} }. 
\end{align}
Since $\sigma^2_{\boldsymbol{i}} \sqrt{N} \frac{\partial \mu}{\partial W_{\boldsymbol{i}}} = \prod_j u_{i_j}$, after multiplying by $\sigma^2_{\boldsymbol{i}}$ and summing over $\boldsymbol{i}$ the first term in the expectation becomes $-N^{-1/2} \scalar{x}{u}/\mu^2$, as all factors $u_{i_j}$ appear twice except for $u_{i_d}$.
Using now the expression for the derivative of $u_{i_j}$, we get 
\begin{align}
 \E{ \frac{1}{\mu} \, \scalar{x}{\tensor{W} \cdot u^{d-1}}} 
   = & \ -\frac{1}{\sqrt{N}} \E{\frac{\scalar{x}{u}}{\mu^2}} \\
   + \sum_{\boldsymbol{i}} & \ \E{ x_{i_d} \, \frac{\sigma^2_{\boldsymbol{i}}}{\mu}
     \sum_{j=1}^{d-1} \left( \prod_{k \neq j}^{d-1} u_{i_k} \right)
     \left[ -\frac{[\bar{R} \phi]_{i_j}}{(d-1)\sqrt{N}}
       + \frac{u_{i_j}^2}{\mu (d-2)\sqrt{N} \sigma^2_{\boldsymbol{i}}} 
         \prod_{m \neq j}^d u_{i_m} \right] }.
\end{align}
The second term inside the inner brackets is easy to compute, since the products of $u_{i_k}$ and of $u_{i_m}$ combine to yield $u_{i_d} \prod_{k \neq j}^{d-1} u_{i_k}^2$, and thus the summation over $\boldsymbol{i}$ leaves only $\scalar{x}{u}$ divided by $\mu^2 (d-2)\sqrt{N}$ for each $j$.
As for the first term, we can expand it as $[\bar{R} \phi]_{i_j} = (d \, \sigma^2_{\boldsymbol{i}})^{-1} \, \sum_{\ell=1}^d \bar{R}_{i_j, i_{\ell}} \prod_{m \neq \ell}^d u_{i_m}$ and note that for $j = \ell$ (there are $d-1$ such terms), we recover after summing over $\boldsymbol{i}$ the a term of the form $-(d \sqrt{N} \mu)^{-1} \scalar{x}{u} \tr \bar{R}$. 
The terms for which $j \neq \ell$ (there are $(d-1)^2$) can be obtained by noting that $\bar{R} u = \frac{d-1}{(d-2)\mu} u$ and by matching the indices of the components of $u$ (with only $u_{i_d}$ remaining free to contract with $x_{i_d}$). This computation gives
\begin{align}
 \E{ \frac{1}{\mu} \, \scalar{x}{\tensor{W} \cdot u^{d-1}}} 
   = & \ -\frac{1}{\sqrt{N}} \, \E{\frac{\scalar{x}{u}}{\mu^2} 
   + \frac{d-1}{(d-2)} \,  \frac{\scalar{x}{u}}{\mu^2} 
   - \frac{1}{d} \,  \frac{\scalar{x}{u}}{\mu} \tr \bar{R} 
   - \frac{(d-1)^2}{d(d-2)} \, \frac{\scalar{x}{u}}{\mu^2} } \\
   = & \ \frac{d - (d-1)^2}{d(d-2)\sqrt{N}} \E{ \frac{\scalar{x}{u}}{\mu^2} }
     - \frac{1}{d\sqrt{N}}\E{ \frac{\scalar{x}{u}}{\mu} \tr \bar{R} }.
\end{align}
We have thus derived the formula
\begin{equation}
\label{exp-align}
  \E{ \scalar{x}{u} }
           = \lambda \, \E{ \frac{1}{\mu} \, \scalar{x}{u}^{d-1} }
            + \frac{d - (d-1)^2}{d(d-2) N} \E{ \frac{\scalar{x}{u}}{\mu^2} }
     - \frac{1}{d N}\E{ \frac{\scalar{x}{u}}{\mu} \tr \bar{R} }.
\end{equation}

\subsection{Concentration of $N^{-1} \, \tr R(z)$ around $m_d(z)$}

Since both expressions \eqref{exp-mu} and \eqref{exp-align} involve the normalized trace of the resolvent $R$ of $\tensor{Y} \cdot u^{d-2}$ evaluated at $\mu/(d-1)$, we will need the asymptotic expression of that quantity. It turns out that this normalized trace converges almost surely to $m_d(z)$, that is, the same Stieltjes transform that was computed for contractions of $\tensor{W}$ with independent unit norm vectors $v$ in Theorem \ref{thm:spec-meas}. We will now show this fact.

From now on, for simplicity of notation we shall omit the argument of $R$ whenever it is simply $z$. As in the previous sections, the notation $\bar{R}$ will stand for $R(\mu/(d-1))$, where $\mu$ is the eigenvalue associated with the vector $u$ that defines the contraction $\tensor{Y} \cdot u^{d-2}$.
Recall that $R$ satisfies the recurrence relation
\begin{equation}
\label{rec-R}
 R = -\frac{1}{z} I + \frac{1}{z}\left( \tensor{Y} \cdot u^{d-2} \right) R.
\end{equation}  
Replacing $\tensor{Y}$ by its definition and taking the expectation of the normalized trace,
\begin{equation}
\label{norm-trace}
 \E{\frac{1}{N} \, \tr R} = -\frac{1}{z} 
  + \frac{\lambda}{N z} \, \E{ \scalar{x}{u}^{d-2} \scalar{x}{R x} }  + \frac{1}{z N \sqrt{N}} \sum_{\boldsymbol{i}} 
      \E{ W_{\boldsymbol{i}} \left( \prod_{k=2}^{d-1} u_{i_k} \right) R_{i_1 i_d} }.
\end{equation}
As usual, Stein's lemma will provide us an expression for the second expectation of the right-hand side:
\begin{equation}
\label{diff-trR}
 \frac{1}{N \sqrt{N}} \sum_{\boldsymbol{i}} \E{ W_{\boldsymbol{i}} \left( \prod_{k=2}^{d-1} u_{i_k} \right) R_{i_1 i_d} }
 = A + B,
\end{equation} 
where 
\begin{align}
 A := & \ \frac{1}{N \sqrt{N}} \sum_{\boldsymbol{i}} \sigma^2_{\boldsymbol{i}} \, 
 \E{\frac{\partial R_{i_1 i_d}}{\partial W_{\boldsymbol{i}}} \left( \prod_{k=2}^{d-1} u_{i_k} \right)}, \\
 B := & \ \frac{1}{N \sqrt{N}} \sum_{\boldsymbol{i}} \sigma^2_{\boldsymbol{i}} \, \E{ R_{i_1 i_d} \sum_{\ell=2}^{d-1} \frac{\partial u_{i_\ell}}{\partial W_{\boldsymbol{i}}} \left( \prod_{k=2, \, k \neq \ell}^{d-1} u_{i_k} \right) }.
\end{align}
In the following, we will compute asymptotic expressions for $A$ and $B$.

\subsubsection{Asymptotic expression of $A$ in \eqref{diff-trR}}

To compute $A$, we will need the derivative of $R$ with respect to each component $\tensor{W}_{\boldsymbol{i}}$.
Starting again from \eqref{rec-R}, it is easy to derive
\begin{equation}
 \frac{\partial R}{\partial W_{\boldsymbol{i}}}
 = -R \, \frac{\partial }{\partial W_{\boldsymbol{i}}} \left( \tensor{Y} \cdot u^{d-2} \right) R,
\end{equation} 
which component-wise reads
\begin{equation}
 \frac{\partial R_{pq}}{\partial W_{\boldsymbol{i}}} 
 = -\frac{1}{\sqrt{N}} \sum_{\boldsymbol{j}} R_{p j_1}
   \left( \frac{\partial W_{\boldsymbol{j}}}{\partial W_{\boldsymbol{i}}} \right)
   \left( \prod_{k=2}^{d-1} u_{j_k} \right) R_{j_d q}
   - \sum_{\boldsymbol{j}} \sum_{\ell=2}^{d-1} R_{p j_1}
     \tensor{Y}_{\boldsymbol{j}} \left( \frac{\partial u_{j_\ell}}{W_{\boldsymbol{i}}} \right) \left(\prod_{k=2, \, k \neq \ell}^{d-1} u_{j_k}\right) R_{j_d q}.
\end{equation} 

Hence, we can write $A = A_1 + A_2$, where
\begin{equation}
 A_1 := -\frac{1}{N^2} \sum_{\boldsymbol{i}} \sum_{\boldsymbol{j}} \sigma^2_{\boldsymbol{i}} \,
   \E{ R_{i_1 j_1}
   \left( \frac{\partial W_{\boldsymbol{j}}}{\partial W_{\boldsymbol{i}}} \right)
   \left( \prod_{k=2}^{d-1} u_{i_k} u_{j_k} \right) R_{j_d i_d} }
\end{equation} 
and
\begin{equation}
\label{def-A2}
 A_2 :=  -\frac{1}{N \sqrt{N}} \, \sum_{\boldsymbol{i}} \sum_{\boldsymbol{j}} \sigma^2_{\boldsymbol{i}} \, 
   \E{ \sum_{\ell=2}^{d-1} R_{i_1 j_1}
     \tensor{Y}_{\boldsymbol{j}} \left( \frac{\partial u_{j_\ell}}{\partial W_{\boldsymbol{i}}} \right) \left(  \prod_{k=2, \, k \neq \ell}^{d-1} u_{i_k} u_{j_k} \right) 
      R_{i_d j_d} u_{i_\ell} }.
\end{equation} 
We focus first on $A_1$.
Since $\frac{\partial W_{\boldsymbol{j}}}{\partial W_{\boldsymbol{i}}} = 1$ if $\boldsymbol{j}$ is a permutation of $\boldsymbol{i}$ and $\frac{\partial W_{\boldsymbol{j}}}{\partial W_{\boldsymbol{i}}} = 0$ otherwise, and since $\sigma^2_{\boldsymbol{i}}$ is given by reciprocal of the number of distinct permutations of (the particular value of) $\boldsymbol{i}$, the value of $A_1$ is proportional to the sum of all $d!$ possible $d$-fold contractions obtained by matching the indices of $\boldsymbol{j}$ with one of the possible permutations of $\boldsymbol{i}$. 
This conclusion is reached by noting that $\sigma^2_{\boldsymbol{i}}$ is precisely equal to $1/d!$ times the number of ``repetitions'' of the term with indices $\boldsymbol{i}$ that are required in order to ``complete'' all such contractions.
Therefore, by taking into account the result of all such possible contractions, we concllude that $A_1$ has the form 
\begin{equation}
\label{def-A}
 A_1 = -\frac{1}{N^2 \, d!} \, \left( K_1 \E{ (\tr R)^2} + K_2 \E{ \tr R^2} 
   + K_3 \E{ \frac{1}{(\mu - z)^2}} + K_4 \E{ \tr R \, \frac{1}{\mu - z}} \right),
\end{equation} 
where the integers $K_i$ count the number of contractions yielding terms of each one of the forms in this expression. In particular, it is easy to see that $K_1 = (d-2)!$, since $(\tr R)^2$ results from each contraction in which $(i_1,j_1)$ and $(i_d,j_d)$ are matched, while all others $d-2$ indices $j_2,\ldots,j_{d-1}$ can be arbitrarily permuted to match $i_2,\ldots,i_{d-1}$.
At this point, we will simplify the subsequent calculations by already droping terms which will vanish at the end in the large-$N$ limit.
We have
\begin{equation}
\label{A-tr2}
  \lim_{N \rightarrow \infty} A_1
   = -\frac{1}{d(d-1)} \, \lim_{N \rightarrow \infty} \E{ \left( \frac{1}{N} \, \tr R \right)^2 },
\end{equation} 
since the terms mutiplied by $K_2$, $K_3$ and $K_4$ are all $o(N^2)$. We will see in a moment that $(\frac{1}{N} \, \tr R)^2 $ will in fact converge almost surely to $(m_d(z))^2$.

The key in the above argument is of course the fact that, in expectation, $\scalar{u}{R \, u} = 1/(\mu - z)$ and $\scalar{u}{u}$ are $O(1)$, while $\tr R$ is $O(N)$.
By the same token, the term $A_2$ defined in \eqref{def-A2} can also be neglected, since the derivative of $u_{j_\ell}$ contains a $1/\sqrt{N}$ factor and the contractions in \eqref{def-A2} only produce $o(N^2)$ terms: all resolvents appearing in the expression are ultimately contracted with one ``copy'' of $u$ (hence no second-order moment $\left( \frac{1}{N} \, \tr R \right)^2$ arises), and contractions of $u$ with $\tensor{Y}$ are also $O(1)$.

Hence, in summary, $A$ is asymptotically given by the expression derived in \eqref{A-tr2}.

\subsubsection{The term $B$ in \eqref{diff-trR} vanishes asymptotically}

Let us proceed by inserting the derivative of $u$ into $B$ to get $B = B_1 + B_2$, where
\begin{align}
 B_1 := & \ \frac{1}{N \sqrt{N}} \sum_{\boldsymbol{i}} \sigma^2_{\boldsymbol{i}} \,
  \E{ R_{i_1 i_d} \left( \prod_{k=2, \, k \neq \ell}^{d-1} u_{i_k} \right)  \sum_{\ell=2}^{d-1} 
        - \frac{1}{(d-1)\sqrt{N}} \frac{1}{d \sigma^2_{\boldsymbol{i}}} \sum_{m=1}^d 
         \bar{R}_{i_\ell, i_m} \left( \prod_{n \neq m}^d u_{i_n} \right) 
         } \\
     = & \ 
     - \frac{1}{N^2 \, d (d-1)} \sum_{\boldsymbol{i}} \, \E{
      \sum_{\ell=2}^{d-1} \sum_{m=1}^d R_{i_1 i_d} \left( \prod_{k=2, \, k \neq \ell}^{d-1} u_{i_k} \right)  
         \bar{R}_{i_\ell, i_m} \left( \prod_{n \neq m}^d u_{i_n} \right) 
         } 
\end{align}
and 
\begin{align}
 B_2 := & \ \frac{1}{N \sqrt{N}} \sum_{\boldsymbol{i}} \sigma^2_{\boldsymbol{i}} \,
  \E{ R_{i_1 i_d} \left( \prod_{k=2, \, k \neq \ell}^{d-1} u_{i_k} \right) \frac{1}{(d-2) \, \mu \, \sigma^2_{\boldsymbol{i}} \, \sqrt{N}} \, \left(  \prod_{j=1}^d u_{i_j} \right)
     \, u_{i_\ell} } \\
     = & \ \frac{1}{N^2 \, (d-2)} \sum_{\boldsymbol{i}}
     \E{\frac{1}{\mu} \, R_{i_1 i_d} \left( \prod_{k=2}^{d-1} u_{i_k} \right) \, \left(  \prod_{j=1}^d u_{i_j} \right) } \\
     = & \ \frac{1}{N^2 \, (d-2)} \E{ \frac{1}{\mu} \, \frac{1}{\mu - z} } \\
     = & \ O(1/N^2).
\end{align}
It is not hard to see that $B_1 = O(1/N)$, since only one trace (that of $\bar{R}$ when $m = \ell$) arises from the contractions. It thus follows that $B = O(1/N)$.

\subsubsection{Concentration and asymptotic expression for the normalized trace}
\label{sec:asymp-trace}

Relying on the same arguments used in the analytic proof of Wigner's semicircle law, it is not difficult to show that, for any $z$ outside the real line, the (analytic) expected normalized trace $\E{ \frac{1}{N} \, \tr R(z) }$ converges to a function of $z$
\begin{equation}
 m_d(z) := \lim_{N \rightarrow \infty} \E{ \frac{1}{N} \, \tr R },  \quad 
 \forall z \in \mathbb{C} \setminus \mathbb{R},
\end{equation} 
which is also analytic by Vitali's theorem.
The first ingredient is the inequality $\norm{R^p(z)} \le 1/|\Im(z)|^p$, valid for all such $z$ and all integers $p$, which entails that all moments $\frac{1}{N} \, \tr R^p$ are bounded on $\mathbb{C} \setminus \mathbb{R}$ with probability one, thus ensuring the existence of convergent subsequences. 
Then, it can be shown that every such subsequence must actually converge to a same limit $m_d(z)$, which by \eqref{norm-trace} must then satisfy 
\begin{equation}
 \label{eq-md}
   m_d(z) = -\frac{1}{z} 
     - \frac{1}{z \, d \, (d-1)} \lim_{N \rightarrow \infty} \E{ \left( \frac{1}{N} \tr R \right)^2 }. 
\end{equation} 
It is now a standard exercise to show that $\frac{1}{N} \, \tr R $ concentrates around $m_d(z)$, and hence the above moment can be replaced by $(m_d(z))^2$. This can be done by resorting to the Nash-Poincaré inequality \cite{PastS-11-book} to bound the variance of the normalized trace as in
\begin{align}
\label{var-norm-tr}
    \text{Var}\left( \frac{1}{N} \, \tr R \right)
     \le \frac{1}{N^2} \, 
     \E{ \sum_{\boldsymbol{i}} 
     \sigma^2_{\boldsymbol{i}} \, \left( \sum_p \frac{\partial R_{pp}}{\partial W_{\boldsymbol{i}}} \right)^2 }. 
\end{align}
Now, denoting the $(i,j)$ component of $R^2$ by $R^2_{ij}$, we can write
\begin{equation}
 \sum_p \frac{\partial R_{pp}}{\partial W_{\boldsymbol{i}}} 
 = -\frac{1}{\sqrt{N}} \sum_{\boldsymbol{j}} R^2_{j_1 j_d}
   \left( \frac{\partial W_{\boldsymbol{j}}}{\partial W_{\boldsymbol{i}}} \right)
   \left( \prod_{k=2}^{d-1} u_{j_k} \right) 
   - \sum_{\boldsymbol{j}} \sum_{m=2}^{d-1} R^2_{j_1 j_d}
     \tensor{Y}_{\boldsymbol{j}} \left( \frac{\partial u_{j_m}}{W_{\boldsymbol{i}}} \right) \left(\prod_{k=2, \, k \neq m}^{d-1} u_{j_k}\right),
\end{equation} 
Using this expression, it is tedious but not hard to verify that the expectation in \eqref{var-norm-tr} is $O(1)$. This esentially comes from the fact that each term in that expectation comprises a factor $1/N$ and at most a single trace (of $R^4$). 
Because all moments $\frac{1}{N} \, \tr R^p$ are bounded on $\mathbb{C}\setminus\mathbb{R}$, it follows that $\text{Var}\left( \frac{1}{N} \tr R(z) \right) = O(1/N^2)$ for $z \in \overline{\mathbb{C}}$, and thus by a standard combination of Markov's inequality and Borel-Cantelli we have that $\frac{1}{N} \tr(R(z)) \convas m_d(z)$ on that set.
Hence, from \eqref{eq-md} we deduce that, as $N \rightarrow \infty$,
\begin{equation}
\label{asymp-md}
 \frac{1}{z \, d(d-1)} (m_d(z))^2 + z \, m_d(z) + 1 = 0, 
\end{equation} 
thus yielding the solutions 
\begin{equation}
 m_d(z) = \frac{-d(d-1) \pm \sqrt{d^2(d-1)^2 z^2 - 4 d (d-1)}}{2}.
\end{equation} 
At last, using our definition $\beta_d = 2/\sqrt{d(d-1)}$ and keeping the solution corresponding to a Stieltjes transform (bevahing as $O(z^{-1})$ at infinity), we arrive at
\begin{equation}
\label{md-proof}
   m_d(z) =  \frac{2}{\beta_d^2} \left( -z + z \sqrt{ 1 - \frac{\beta_d^2}{z^2} } \right),
\end{equation} 
which is the same expression of Theorem \ref{thm:spec-meas}.
Hence, by the Stieltjes continuity theorem (see Appendix \ref{app:RMT-tools}), the spectrum of $\tensor{Y} \cdot u^{d-2}$ converges weakly almost surely to a semi-circle law supported on $[-\beta_d, \beta_d]$.

Let us pause for a moment to reflect on the implications of this intermediate result. Since the almost sure limit $\mu_{d,\infty}(\lambda)$ of $\mu$ obtained by our calculations must be such that $\mu_{d,\infty}(\lambda)/(d-1)$ is not in the limiting spectrum of $\tensor{Y} \cdot u^{d-2}$ (because of the factor $R(\mu/(d-1))$ in the formulas for the derivatives of $\mu$ and $u$), and since the limiting spectral measure of this matrix is a semicircle law supported on $[-\beta_d,\beta_d]$, it follows that we must have $\mu_{d,\infty}(\lambda) > (d-1) \beta_d$ for our formulas to make sense---just as we assume in the statement of Theorem \ref{thm:main}.

Incidentally, the observation of Remark \ref{rem:mu}, combined with the limiting spectrum of $\tensor{Y} \cdot u^{d-2}$, implies that the existence of local maxima associated with negative eigenvalues $\mu$ becomes increasingly unlikely, since the symmetric support of the limiting spectral measure $[-\beta_d,\beta_d]$ is incompatible with the condition that all other eigenvalues must lie in $]-\infty, \mu/(d-1)]$.
This observation is coherent with the results of \cite{ArouMMN-19-CPAM}, which show that the expected number of local maxima associated to negative eigenvalues decays exponentially as $N \rightarrow 0$.

\subsubsection{A momentary digression: extending the above proof to show Theorem \ref{thm:spec-meas}}
\label{sec:spec-meas}

The proof of Theorem \ref{thm:main} can be obtained by adapting the above derivation. The idea is to set $\lambda = 0$ and replace $u$ by a determinstic vector $v \in \mathbb{S}^{N-1}$ in \eqref{rec-R}:
\begin{equation}
  Q = -\frac{1}{z} I + \frac{1}{z \, \sqrt{N}} (\tensor{W} \cdot v^{d-2}) \, Q,
\end{equation}
where we use $Q$ to denote this new resolvent to avoid any ambiguities. We are as before interested in the expected normalized trace of $Q$,
\begin{equation}
\label{norm-trace-Q}
 \E{\frac{1}{N} \, \tr Q} = -\frac{1}{z} 
  + \frac{1}{z N \sqrt{N}} \sum_{\boldsymbol{i}} 
      \E{ W_{\boldsymbol{i}} \left( \prod_{k=2}^{d-1} v_{i_k} \right) Q_{i_1 i_d} }.
\end{equation}
Since now $v$ does not depend on $\tensor{W}$, the calculations are greatly simplified: the derivative of $Q$ is
\begin{equation}
 \frac{\partial Q_{pq}}{\partial W_{\boldsymbol{i}}} 
 = -\frac{1}{\sqrt{N}} \sum_{\boldsymbol{j}} Q_{p j_1}
   \left( \frac{\partial W_{\boldsymbol{j}}}{\partial W_{\boldsymbol{i}}} \right)
   \left( \prod_{k=2}^{d-1} v_{j_k} \right) Q_{j_d q},
\end{equation} 
and therefore the application of Stein's lemma to compute the expectation in the right-hand side of \eqref{norm-trace-Q} gives
\begin{equation}
 \sum_{\boldsymbol{i}} \E{ W_{\boldsymbol{i}} \left( \prod_{k=2}^{d-1} v_{i_k} \right) Q_{i_1 i_d} }
 = -\frac{1}{\sqrt{N}} \sum_{\boldsymbol{i}} \sum_{\boldsymbol{j}} \sigma^2_{\boldsymbol{i}} \,
   \E{ Q_{i_1 j_1}
   \left( \frac{\partial W_{\boldsymbol{j}}}{\partial W_{\boldsymbol{i}}} \right)
   \left( \prod_{k=2}^{d-1} v_{i_k} v_{j_k} \right) Q_{j_d i_d} }.
\end{equation} 
Here again, by considering all possible matchings of the indices $\boldsymbol{i}$ and $\boldsymbol{j}$, we conclude that 
\begin{equation}
 \lim_{N \rightarrow \infty}
 \frac{1}{N \sqrt{N}} \sum_{\boldsymbol{i}} \E{ W_{\boldsymbol{i}} \left( \prod_{k=2}^{d-1} v_{i_k} \right) Q_{i_1 i_d} }
 = - \frac{1}{d(d-1)} \lim_{N \rightarrow \infty} \E{\left( \frac{1}{N} \tr Q \right)^2}, 
\end{equation} 
since all other terms are $o(1)$. In particular, contractions of $Q$ with $v$ are $O(1)$ in expectation, since by Cauchy-Schwarz $|\scalar{Qv}{v}| \le \norm{Qv} \, \norm{v} \le \norm{Q} \le 1/|\Im(z)|$ for all $z \in \mathbb{C}\setminus\mathbb{R}$. We thus get the same expression as in the study of $R$.

Finally, the same reasoning followed in the previous subsection to show concentration of $\frac{1}{N} \tr R$ can be mimicked to show concentration of $\frac{1}{N} \tr Q$, thus leading to the same asymptotic equation as before, namely, to \eqref{asymp-md}. This shows that the spectrum of $\frac{1}{\sqrt{N}} \, \tensor{W} \cdot v^{d-2}$ also converges weakly almost surely to a semicircle law supported on $[-\beta_d, \beta_d]$, as claimed. In retrospect, we can see that the additional terms that are present in the computation of $\frac{1}{N} \tr R$ (originating from the dependence of $u$ on $\tensor{W}$) are all $o(1)$, and thus the resulting spectral measures are the same in both cases.

\subsection{Asymptotic fixed-point equation}

We turn back now to the proof of Theorem \ref{thm:main}, by invoking the key assumption that\footnote{The reader will note that this assumption implicitly imposes that $\lim_{N \rightarrow \infty} \frac{1}{N} \tr R = m_d(z)$ not only outside the real line, but also outside the support $[-\beta_d, \beta_d]$, since the formulas \eqref{exp-mu}--\eqref{exp-align} both involve $m_d$ evaluated on the \emph{real} (random) quantity $\mu/(d-1)$, which by hypothesis converges to a number $\mu_{d,\infty}$ outside the support.}
\begin{align}
   \mu \convas & \ \mu_{d, \infty}(\lambda)  > (d-1) \beta_d,  \\
   \scalar{x}{u} \convas & \ \alpha_{d, \infty}(\lambda) > 0
\end{align}
as $N \rightarrow \infty$. In particular, the first above statement guarantees that the derived derivative formulas hold as $N \rightarrow \infty$, since we know from the previous subsection that the spectrum of $\tensor{Y} \cdot u^{d-2}$ converges to a semi-circle law on $[-\beta_d, \beta_d]$, and hence $\mu_{d, \infty}(\lambda)/(d-1)$ does not fall on this support. 
Hence, by inspection of \eqref{exp-mu}--\eqref{exp-align} we have 
\begin{align}
     \label{alpha-d-proof}
   \alpha_{d, \infty}(\lambda) = & \ \frac{\lambda}{\mu_{d, \infty}(\lambda)}
      \, (\alpha_{d, \infty}(\lambda))^{d-1} - \frac{1}{d} \, \frac{\alpha_{d, \infty}(\lambda)}{\mu_{d, \infty}(\lambda)} \, m_d\left( \frac{\mu_{d, \infty}(\lambda)}{d-1} \right) \\
    \mu_{d, \infty}(\lambda) = & \ \lambda \, (\alpha_{d, \infty}(\lambda))^{d}
     - \frac{1}{d-1} m_d\left( \frac{\mu_{d, \infty}(\lambda)}{d-1} \right).
     \label{mu-d-proof}
\end{align}
Since $\alpha_{d, \infty}(\lambda) > 0$, we get from \eqref{alpha-d-proof}
\begin{equation}
 \mu_{d, \infty}(\lambda) = \lambda
      \, (\alpha_{d, \infty}(\lambda))^{d-2} - \frac{1}{d} \, m_d\left( \frac{\mu_{d, \infty}(\lambda)}{d-1} \right) 
\end{equation}
Solving now for $\alpha_{d, \infty}(\lambda)$, we get 
\begin{equation}
\alpha_{d, \infty}(\lambda) = 
 \left[ \frac{1}{\lambda} \left( \, \mu_{d, \infty}(\lambda) + \frac{1}{d} \, m_d\left( \frac{\mu_{d, \infty}(\lambda)}{d-1} \right) \right)  \right]^{\frac{1}{d-2}}.
\end{equation} 
The equations \eqref{fp-1}--\eqref{fp-2} are thus established, and the proof is complete.

%%%%%%%%%%%%%%%%%%%%%%%%%%%%%%%%%%%%%%%%%%%
%%%%%%%%%%%%%%%%%%%%%%%%%%%%%%%%%%%%%%%%%%%
%%%%%%%%%%%%%%%%%%%%%%%%%%%%%%%%%%%%%%%%%%%
%%%%%%%%%%%%%%%%%%%%%%%%%%%%%%%%%%%%%%%%%%%
%%%%%%%%%%%%%%%%%%%%%%%%%%%%%%%%%%%%%%%%%%%
%%%%%%%%%%%%%%%%%%%%%%%%%%%%%%%%%%%%%%%%%%%
%%%%%%%%%%%%%%%%%%%%%%%%%%%%%%%%%%%%%%%%%%%
%%%%%%%%%%%%%%%%%%%%%%%%%%%%%%%%%%%%%%%%%%%

% --------------------------------------------------------------------------

\section*{Acknowledgements}

This work was supported by the MIAI LargeDATA Chair at University Grenoble Alpes led by R.~Couillet, and also by the Labex CIMI project of the first author.
We would like to thank Gérard Ben-Arous, Mohamed Seddik, Yang Qi and Arthur Marmin for helpful discussions on the subject.
Finally, the anonymous Referees are also thanked for their constructive suggestions which led to significant improvements in the manuscript.

\appendix 

\begin{appendices}

% ==========================================

\section{Derivative of tensor eigenpairs}
\label{app:deriv-eig}

Given an eigenpair $(\mu,u)$, the derivatives of $\mu$ and $u$ with respect to any entry $W_{i_i \, \dots \, i_d}$ can be computed by resorting to the implicit value theorem (see, e.g., \cite{KranP-13-book}). To this end, we start from the system of equations that characterizes the eigenpair $(\mu,u)$:
\begin{align}
  F(\mu, u, \tensor{W}) = 
  \begin{pmatrix}
     \tensor{Y} \, u^{d-1} - \mu \, u \\[2mm]
     1 - \scalar{u}{u}
  \end{pmatrix}
 = 0.
\end{align}
The Jacobian of $F$ with respect to $(\mu,u)$ is
\begin{equation}
 J(\mu, u, \tensor{W}) = 
 \begin{pmatrix}
    (d-1) \tensor{Y} \, u^{d-2} - \mu I  & -u\\[2mm]
      - 2 u^\T & 0.
 \end{pmatrix}
 \end{equation}
 Now, if $\frac{\mu}{d-1}$ is not in the spectrum of $\tensor{Y} \, u^{d-2}$, then the first block of the above Jacobian is invertible, and its inverse is simply $(d-1) R\left( \frac{\mu}{d-1} \right)$, where $R(z) = \left( \tensor{Y} \, u^{d-2} - z I  \right)^{-1}$.
 For simplicity of notation, we will write $\bar{R} := R\left( \frac{\mu}{d-1} \right)$ in the following. In particular, this implies
\[
  \bar{R} u = \frac{1}{\mu - \frac{\mu}{d-1}} \, u = \frac{d-1}{(d-2) \, \mu} \, u.
\]
Defining also
\[
 A := (d-1) \, \bar{R}^{-1}, \quad
 B := -u,  \quad \text{and} \quad
 C := -2 u^\T 
\]
and assuming that $\frac{\mu}{(d-1)}$ is not in the spectrum of $\tensor{Y} \, u^{d-2}$, the Schur complement formula yields
\[
  J^{-1}(\mu, u, \tensor{W}) = 
  \begin{pmatrix}
     A^{-1} + A^{-1} B S^{-1} C A^{-1} & -A^{-1} B S^{-1}  \\[2mm]
     -S^{-1} C A^{-1} & S^{-1},
  \end{pmatrix}
\]
where 
\[
 S = - C A^{-1} B = -\frac{2}{d-1} u^T \bar{R} u 
                  = -\frac{2}{(d-2) \, \mu}.
\]
Hence,
\begin{equation}
J^{-1}(\mu, u, \tensor{W}) = 
  \begin{pmatrix}
     \frac{1}{d-1}\bar{R} - \frac{1}{(d-2) \, \mu}  u u^\T 
     & 
     -\frac{1}{2} \, u \\
     -u^T &
     - \frac{d-2}{2} \, \mu
  \end{pmatrix}.
\end{equation} 

Now, by the implicit function theorem we have 
\[
  \begin{pmatrix}
    \displaystyle \frac{\partial u}{\partial W_{i_1 \dots i_d}} \\[4mm]
    \displaystyle \frac{\partial \mu}{\partial W_{i_1 \dots i_d}}
  \end{pmatrix}
  = -J^{-1}(\mu, u, \tensor{W}) \, \frac{\partial F}{\partial W_{i_1 \dots i_d}}(\mu, u, \tensor{W}).
\]
The required derivative $\frac{\partial F}{\partial W_{i_1 \dots i_d}}$ turns out to be 
\[
 \frac{\partial F}{\partial W_{i_1 \dots i_d}}(\mu, u, \tensor{W})
 = \begin{pmatrix}
      \frac{1}{\sqrt{N}} \, \phi \\ 0
   \end{pmatrix},
\]
where $\phi$ is an $N$-dimensional vector with at most $d$ nonzero entries given by
\begin{equation}
 \label{phi-vec}
  \sigma^2_{W_{i_1 \dots i_d}} \, \phi
  = \frac{1}{d} \, \sum_{j=1}^d u_1 \dots u_{i_{j-1}} \, u_{i_{j+1}} \dots u_d \, e^{(i_j)}.
\end{equation} 
Hence, provided that $\frac{\mu}{d-1}$ is not an eigenvalue of $\tensor{Y} \, u^{d-2}$, we finally get the derivatives
\begin{equation}
\label{deriv-eigen}
  \begin{pmatrix}
    \displaystyle \frac{\partial u}{\partial W_{i_1 \dots i_d}} \\[4mm]
    \displaystyle \frac{\partial \mu}{\partial W_{i_1 \dots i_d}}
  \end{pmatrix}
  = 
  \begin{pmatrix}
  \displaystyle
  - \frac{1}{(d-1)\sqrt{N}}\bar{R} \, \phi + \frac{1}{(d-2) \, \mu} \, \frac{\partial \mu}{\partial W_{i_1 \dots i_d}} \, u
 \\[4mm]
   \displaystyle \frac{1}{\sigma^2_{w_{i_1 \dots i_d}} \, \sqrt{N}} \, \prod_{j=1}^d u_{i_j}
  \end{pmatrix}.
\end{equation} 

For instance, for $d=3$ we have 
\[
  \begin{pmatrix}
    \displaystyle \frac{\partial u}{\partial W_{i m \ell}} \\[4mm]
    \displaystyle \frac{\partial \mu}{\partial W_{i m \ell}}
  \end{pmatrix}
  = 
  \begin{pmatrix}
  \displaystyle
  - \frac{1}{2\sqrt{N}}\bar{R} \, \phi + \frac{1}{\mu} \, \frac{\partial \mu}{\partial W_{i m \ell}} \, u
 \\[4mm]
   \displaystyle \frac{1}{\sigma^2_{W_{im \ell}} \sqrt{N}} \, u_i u_m u_\ell
  \end{pmatrix}.
\]

% ====================================================================================

\section{Proof of Lemma \ref{lem:cont-diff}}
\label{app:cont-diff}

For every triple $T = (\mu, u, \tensor{W})$ such that the Jacobian $J(\mu, u, \tensor{W})$ of $F$ (see Appendix \ref{app:deriv-eig}) with respect to $(\mu, u)$ is nonsingular, by the implicit function theorem there exists a neighborhood of $T$ such that $(\mu, u)$ is given by a (unique) continuously differentiable function of $\tensor{W}$ whose derivative satisfies \eqref{deriv-eigen}. 
Hence, provided that the set 
$$ B = \{\tensor{W} \in \mathcal{S}^d(N)\ : \ J(\mu, u, \tensor{W}) \text{ is singular } \forall (\mu, u) \text { satisfying } F(\mu, u, \tensor{W}) = 0 \}$$ 
has measure zero, then for almost every $\tensor{W} \in \mathcal{S}^d(N)$ there exists at least one couple $(\mu, u)$ such that the theorem can be applied to $(\mu, u, \tensor{W})$.
It follows that one can construct (at least) one function $G : \mathcal{S}^d(N) \rightarrow \mathbb{R}^{N+1}$ defined and continuously differentiable almost everywhere on $\mathcal{S}^d(N)$ such that $G(\tensor{W}) = (\mu, u)$ is an eigenpair of $\tensor{W}$, with $u \in \mathbb{S}^{N-1}$. 

Hence, we are left with the proof that the above set $B$ indeed has measure zero. In particular, every tensor $\tensor{W} \in \mathcal{S}^d(N)$ having at least one eigenpair $(\mu, u)$ such that $\mu/(d-1)$ is not an eigenvalue of the contraction $\tensor{W} \cdot u^{d-2}$ cannot belong to $B$, since the resolvent of such contraction is well defined on $\mu/(d-1)$, and hence we can apply the explicit expression (in terms of that resolvent) for the inverse of $J(\mu, u, \tensor{W})$ given in Appendix \ref{app:deriv-eig}. But it turns out that the set 
$$B' = \{ \tensor{W} \in \mathcal{S}^d(N) \ : \ \tensor{W} \text{ has one eigenpair } (\mu, u) \text{ such that } \mu/(d-1) \text{ is an eigenvalue of } \tensor{W} \cdot u^{d-2}\},$$
which contains $B$, has measure zero. To show this, it suffices to construct an example of a tensor $\tensor{W}_0$ not in $B'$, since this implies the existence of a non-empty open set of tensors $U$ (a neighborhood of $\tensor{W}_0$) such that $U \cap B' = \emptyset$. In other words, $B'$ would then be proper Zariski closed in $\mathcal{S}^d(N)$, and hence of zero measure.
Such an example can be easily constructed. For instance, take any rank-one tensor $v^{\otimes d}$ with $v \in \mathbb{S}^{N-1}$: $(1,v)$ is clearly an eigenpair of $\tensor{W}$, but $1/(d-1)$ is not an eigenvalue of $v \otimes v$.

% ====================================================================================

\section{Main tools in Pastur's Stein approach}
\label{app:RMT-tools}

We describe in the following the main tools used in the so-called Pastur's Stein approach for the study of random matrices, which mainly revolve around the resolvent formalism. In this description, we will denote the $N$ eigenvalues of a symmetric matrix $Y \in \mathbb{R}^{N \times N}$ by $\mu_1(Y) \le \dots \le \mu_N(Y)$, and its spectral decomposition by
\begin{equation}
    Y = \sum_{i=1}^N \mu_i(Y) \, u_i(Y) \,  u_i^\T(Y),
\end{equation}
where the dependence of the eigenvalues and eigenvectors on $Y$ will be omitted whenever no ambiguity arises, for the sake of simplicity.

\subsection{The resolvent matrix and its properties}

Let us start by defining the resolvent of a symmetric matrix.

\begin{dfn}[Resolvent]
 Given a symmetric matrix $Y \in \mathbb{R}^{N \times N}$, we define its resolvent matrix as 
 \[
   R_Y(z) := (Y - z I)^{-1}, \qquad z \in \mathbb{C}\setminus\sigma(Y), 
 \]
 where $\sigma(Y)$ stands for the spectrum of $Y$.
\end{dfn}

The resolvent is a rich object that encodes spectral properties of $Y$ in its complex-analytic structure.
Among its various properties, we can mention the following:
\begin{enumerate}[(i)]
    \item it satisfies the recurrence relations
\begin{align}
\label{res-rec}
 R_Y(z) = & \ -\frac{1}{z} I + \frac{1}{z}  \, Y R_Y(z) 
      = -\frac{1}{z} I + \frac{1}{z} \, R_Y(z) Y;
% \label{Q-prop-2}      
\end{align}
    
    \item \label{prop:spec} its spectral decomposition $R_Y(z) = \sum_{i=1}^N \frac{1}{\mu_i - z} \, u_i \,  u_i^\T$ is a meromorphic function with poles on the eigenvalues of $Y$;
    
    \item its trace $\tr R(z) = \sum_{i=1}^N \frac{1}{\mu_i - z}$ is a complex analytic function on $\mathbb{C} \setminus \sigma(Y)$;
        
    \item \label{prop:proj} as a consequence of (\ref{prop:spec}), the contour integral 
    \[
      P_{\mu_i} = - \frac{1}{2\pi i} \oint_{C_{\mu_i}} R_N(z) \, dz
    \]
    yields a projector onto the eigenspace associated with $\mu_i$, where $C_{\mu_i}$ denotes a sufficiently small positively oriented contour around eigenvalue $\mu_i$ alone.

\end{enumerate}

Property \eqref{prop:proj} is often utilized to get the alignment of an arbitrary vector $x$ with the eigenvector $u$ of an eigenvalue $\mu$ of multiplicity one, via:
\begin{equation}
\label{align-formula}
    \langle u, x \rangle^2 = 
    x^\T \, P_{\mu} \, x = - \frac{1}{2\pi i} \oint_{C_{\mu}} x^\T \, R_Y(z) \, x \, dz.
\end{equation}
In particular, when studying a sequence of random $N \times N$ matrices of growing dimension $N$, the above formula is useful for computing the asymptotic alignment of a given sequence of vectors $x$ with the sequence of eigenvectors $u$ whose eigenvalues converge to a limit $\mu$ which is bounded away from the support of the (rest of the) limiting spectrum.

\subsection{The Stieltjes transform}

One typical question that arises in the study of random matrices is whether the empirical spectral distribution 
\begin{equation}
\label{empi-meas}
  \rho_{Y_N}  \, := \, \frac{1}{N} \, \sum_{i=1}^N \delta_{\mu_i(Y_N)},
\end{equation}
where $\delta_{\mu}$ is a Dirac mass on $\mu$, of a sequence $\{Y_N\}$ of random $N \times N$ matrices converges to some deterministic probability measure $\rho$. One widely used method that allows answering this question is based upon an analytic tool known as the Stieltjes transform, defined next.

\begin{dfn}[Stieltjes transform]
Given a probability measure $\rho$, its Stieltjes transform is defined by
\begin{equation}
    s_{\rho}(z) := \int \frac{1}{t - z} \, \rho(dt), \qquad z \in \mathbb{C} \setminus \text{supp}(\rho).
\end{equation}
\end{dfn}
The usefulness of this tool in RMT is explained by the following result, discussed e.g.~by \cite{Tao-12-book} (see Exercise 1.1.25 of that book).

\begin{thrm}[Stieltjes' continuity theorem]
The sequence of random probability measures $\rho_k$ supported on $\mathbb R$ converges almost surely weakly\footnote{While this result is often stated in terms of convergence in the vague topology rather than weak topology, these are equivalent when the limiting measure $\rho$ is assumed to be a probability measure, since no mass is lost.} to the deterministic measure $\rho$ if and only if $s_{\rho_k}(z) \convas s_{\rho}(z)$ for all complex $z$ in the upper-half plane $\overline{\mathbb{C}} := \mathbb{R} + i \, \mathbb{R}_+$.
\end{thrm}

It should be noted that the same result holds also for other forms of convergence apart from almost sure (for instance, convergence in probability).
In any case, the above posed question of convergence of $\rho_{Y_N} \overset{?}{\rightarrow} \rho$ can be translated into a question of pointwise convergence of a sequence of Stieltjes transforms. Furthermore, if $\rho$ has a density $f$, then having determined the limiting transform $s_{\rho}(z)$ this density can be recovered by the following identity (often termed Sokhotski--Plemelj formula): 
\begin{equation}
  f(x) = 
    \frac{1}{\pi} \, 
    \lim_{\epsilon \rightarrow 0^+} 
      \Im\{ s_{\rho}(x + i \epsilon) \};
\end{equation}
see, e.g., \citep{Tao-12-book}.

In the particular case of a sequence of probability measures \eqref{empi-meas}, it can be seen that
\[
 s_{\rho_{Y_N}}(z) = \frac{1}{N} \, \sum_{i=1}^N \frac{1}{\mu_i - z}
                   = \frac{1}{N} \, \tr R_{Y_N}(z),
\]
which reinforces the central role played by the resolvent formalism.

\subsection{Gaussian integration by parts (Stein's lemma)}

The last technical tool that we will review is Stein's lemma, also known as Stein's identity or Gaussian integration by parts, which allows replacing the expected product of a Gaussian variable with a differentiable function $f$ (of polynomially bounded growth) by the variance of that variable times the expectation of $f'$.

\begin{lem}[\cite{Stei-81-AS}]
If $x \sim \mathcal{N}(0,\sigma^2)$ and $f : \mathbb{R} \rightarrow \mathbb{R}$ is continuously differentiable almost everywhere and has at most polynomial growth, then
\[
 \mathbb{E}\{x \, f(x)\} = \sigma^2 \, \mathbb{E}\{f'(x)\}.
\]
\end{lem}

% ====================================================================================

\section{Maple solution of fixed-point equation for $d=3$}
\label{app:maple}

By simply defining on Maple: 
\begin{align}
  & \text{\tt h(z) := sqrt(z\^{}2 - 2/3)} \\
  & \text{\tt omega(z) := (1/2*z + h(1/2*z))/lambda } \\
  & \text{\tt phi(z) := lambda*omega(z)\^{}3 + (3/4)*z - (3/2)*h(z/2)} 
\end{align}
(note that $m_3(z) = -3z + 3h(z)$ with $h(z) = \sqrt{z^2 - 2/3}$), one can find the solutions of $z = \phi(z, \lambda)$ for the fixed-point equation in the case $d=3$ by using the command:
\[
   \text{\tt solve(z = phi(z), {z})}.
\]
This produced four solutions:
\begin{align}
\label{soln-maple}
  z_{1,2}^\star(\lambda)=\frac{\sqrt{18 \lambda^{4}+72 \lambda^{2}
  \pm 2\sqrt{3}\, \sqrt{\lambda^{2} \left(3 \lambda^{2}-4\right)^{3}}}}{6 \lambda},
  \quad
  z_3^\star(\lambda) = -z_1^\star(\lambda) 
  \quad \text{and} \quad
  z_4^\star(\lambda) = -z_2^\star(\lambda).
\end{align}
We now show that the solution $z_1^\star(\lambda)$ corresponds precisely to the formula \eqref{mu-JagaML} for the case $\lambda > \lambc(d)$.
As for the other solutions, $z_3^\star$ and $z_4^\star$ are negative and thus not of our interest, while $z_2^\star$ holds only for negative values of $\lambda$. 
Letting $\theta=\lambda \sqrt{3}$, we get
\begin{equation}
    \mu^\star_3(\lambda) = \frac{1}{\sqrt{6}}\frac{\theta^2 + \theta\sqrt{\theta^2 - 4}+4}{\sqrt{\theta^2 + \theta\sqrt{\theta^2 - 4}}}
\quad \text{and} \quad
    z_1^\star(\lambda) = \frac{1}{\sqrt{6}}\sqrt{\theta^2+12+\frac{1}{\theta}(\theta^2-4)^{3/2}}.
\end{equation}
Defining also $\omega=\sqrt{\theta^2-4}=\sqrt{3\lambda^2-4}$, we can write the squares of these functions as
\[
[\mu_3^\star(\lambda)]^2 = \frac{1}{24 \, \theta}(\theta^2+\theta \omega+4)^2(\theta-\omega)
\]
and
\begin{equation}\label{zm}
 [z_1^\star(\lambda)]^2 = \frac{1}{6 \, \theta}(\theta^3+\theta^2 \omega + 12 \theta -4 \omega).
\end{equation}
Finally, by developing the above expressions and replacing $\omega^2$ by $(\theta^2-4)$ to obtain a polynomial in $(\theta,\omega)$ of partial degree 1 in $\omega$, one can easily show that $[\mu_3^\star(\lambda)]^2 = [z_1^\star(\lambda)]^2$. Since both formulas yield positive values, this implies $\mu_3^\star(\lambda) = z_1^\star(\lambda)$, as claimed.

\end{appendices}

\bibliographystyle{plainnat}
\bibliography{main.bib}

\end{document}